%% file: draft.tex
\documentclass[sigconf, natbib=false, nonacm]{acmart}

\usepackage[datamodel=acmdatamodel, style=acmnumeric, sorting=none, backend=biber]{biblatex}
\usepackage{amsmath, amsfonts}

\usepackage{orcidlink}

\usepackage{graphicx}
\usepackage{subcaption}

\usepackage[inline]{enumitem}
\setlist*[itemize]{labelindent=10pt, itemindent=0pt, leftmargin=*}

\usepackage{booktabs}
\usepackage{multirow}
\usepackage{tabularx}

\usepackage{pgfplots}
\pgfplotsset{compat=1.18}
\usepgfplotslibrary{statistics}

\usepackage[nameinlink]{cleveref}

\usepackage{csquotes}
\usepackage{textcomp}

\usepackage{balance}

\begin{document}

\title[Transformer Inference Deployment on FPGA: Survey]{Recent Developments in Transformer Inference Deployment on FPGA Platforms: A Survey}

\author{Arjan {Blankestijn}}
\orcid{0009-0005-4628-7689}
\affiliation{%
	\institution{Computer Architecture for Embedded Systems, University of Twente}
	\city{Enschede}
	\country{The Netherlands}}
\email{a.blankestijn-2@student.utwente.nl}

\author{Uraz {Odyurt}}
\orcid{0000-0003-1094-0234}
\affiliation{%
	\institution{Faculty of Engineering Technology, University of Twente}
	\city{Enschede}
	\country{The Netherlands}}
\email{u.odyurt@utwente.nl}

\author{Amirreza {Yousefzadeh}}
\orcid{0000-0002-2967-5090}
\affiliation{%
	\institution{Computer Architecture for Embedded Systems, University of Twente}
	\city{Enschede}
	\country{The Netherlands}}
\email{a.yousefzadeh@utwente.nl}

\renewcommand{\shortauthors}{A. Blankestijn et al.}

\begin{abstract}
With the rapid and continuous growth in the incorporation of machine learning models based on the Transformer architecture, capable deployment is in high demand. In this context, capable deployment refers to operational performance aspects, e.g., throughput and latency, as well as efficiency aspects, e.g., energy consumption. When it comes to the task of inference using such models, purpose-built hardware accelerators provide a lucrative alternative to common deployment choices, such as Central Processing Units (CPUs) and Graphics Processing Units (GPUs). The Field Programmable Gate Array (FPGA) platforms category is an example of such alternative accelerators, promising implementation flexibility, energy efficiency, improved latency and suitability for on-site deployment. We investigate the most recent advances, trends, and design choices for Transformer inference on FPGA platforms. We perform a systematic literature review, extracting and delving into preferred techniques for implementation and optimisation. This study and the provided taxonomy of topics could act as a guide for researchers from the academia and industry alike.

\end{abstract}

%


\maketitle

\input{text/body}

\balance

\printbibliography

\end{document}

%% file: text/body.tex

\section{Introduction}
\label{sec:introduction}
Machine Learning (ML) has dramatically reshaped the landscape of computational algorithm design in numerous contexts. The inherent data parallelism present in ML models has proven to be highly advantageous. Additionally, specialised hardware accelerator architectures have been built and incorporated, increasing the computational performance of ML models even further. Well-known examples of such specialised hardware architectures beyond Central Processing Units (CPUs) are, Graphics Processing Units (GPUs), Neural Processing Units (NPUs), Field-Programmable Gate Arrays (FPGAs), and Neuromorphic computing. Some of these designs pre-existed and found to be effective ML accelerators for different reasons, i.e., inherent computational compatibilities or fine-grained customisability. Two main examples are GPUs and FPGAs.

Amongst these hardware accelerator architectures, FPGAs stand out by portraying unique characteristics. FPGAs have attracted substantial attention as a viable architecture for accelerating deep learning workloads. Their reconfigurability, together with the capacity for custom, hardware-level optimisations, makes FPGAs particularly attractive for ML inference. Example use-cases benefiting from hardware-level optimisations are requirements for low-latency operation, high-throughput operation, or any combination with desired balance between latency and throughput. FPGA-based platforms present a compelling alternative to conventional CPUs and GPU accelerators, especially where power footprint and physical size requirements impose limitations, or where on-site deployment and installation is required.

While established deep learning ML model architectures, e.g., Convolutional Neural Networks (CNNs), have been taking advantage of hardware accelerators, particularly FPGAs, there is an emerging architecture, the Transformer. The Transformer model architecture~\cite{Vaswani:2017:Attention} has gained prominence due to its powerful ability to handle distant and obscure dependencies and to efficiently parallelise computations. Originally conceived for tasks such as machine translation, the Transformer architecture has evolved into a fundamental component for many high-performance models. However, the computational load and memory demands associated with Transformer inference remain significant, with much higher demands compared to the established model architectures. These challenges are especially apparent when Transformer models are intended for real-time applications and deployed in resource-constrained environments.

In short, as opposed to older, more established ML model architectures, the methodology and tooling for deployment of Transformers on FPGAs remain incomplete. We strive to capture the current state-of-the-art for the context of \emph{Transformer inference running on FPGA platforms}. As such, answers to a number of questions are to be sought after:
\begin{itemize}
	\item \emph{What are the performance indicators?}
	\item \emph{How to improve these performance indicators?}
	\item \emph{Which implementation or resource management aspects are to be leveraged to achieve better results?}
	\item \emph{Which optimisations are considered? Which tooling is available and what are the limitations of such tooling?}
\end{itemize}

\paragraph*{Scope of the survey}
We provide a systematic literature review, aiming to consolidate and critically evaluate existing research on Transformer model inference on FPGAs. It must be emphasised that the focus is \emph{inference} and not training. As this particular angle is rather active, we focus on the more recent research from the last 2 years, 2024--May 2025. This focus is deliberate. Rather than aiming to provide a broad historical survey of FPGA-based Transformer acceleration, we seek to capture a focused snapshot of the most recent developments, in which implementation practices, optimisation strategies, and target workloads are evolving rapidly. In this sense, the present survey is intended to complement earlier broader reviews by emphasising recent trends and newly reported design choices.

Following this introduction, the background and motivation, covering alternative surveys and fundamental topics, is provided in \Cref{sec:background}. The survey methodology itself is given in \Cref{sec:survey_method}, followed by the extracted taxonomy of topics in \Cref{sec:taxonomy}. Details on storage, tooling, model compression, and IP design from the covered articles are provided in \Cref{sec:model_weights_storage,sec:incorporated_tooling,sec:model_compression,sec:ip_design_optimisation}, respectively. Analysis of performance metrics and trends is elaborated in \Cref{sec:insights}, alongside concluding remarks in \Cref{sec:conclusion}.

\section{Background and motivation}
\label{sec:background}
\begin{figure}[htbp]
	\centering
	\includegraphics[width=\linewidth]{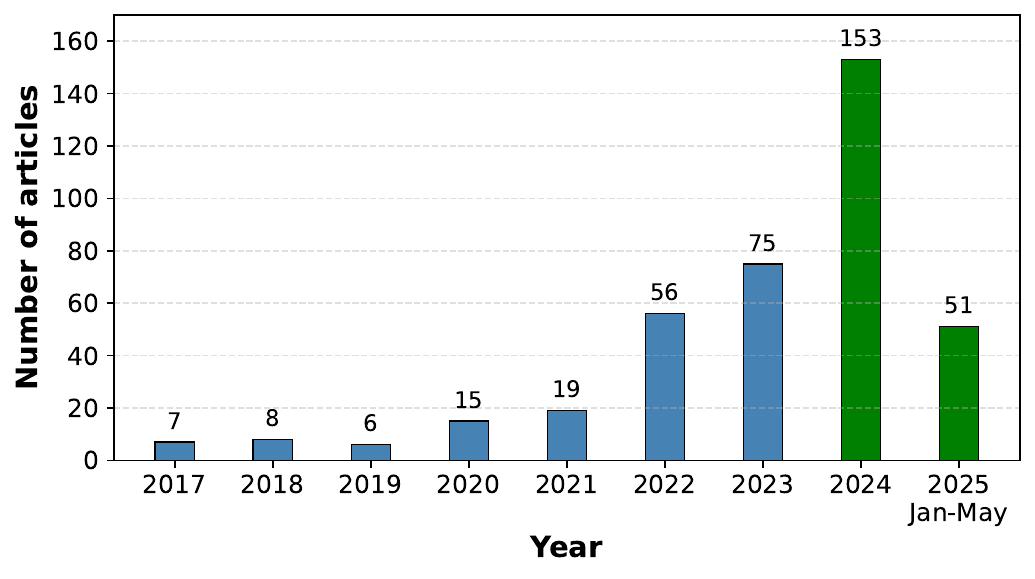}
	\caption{Distribution of article counts per year (till the end of May 2025), focusing on the topic of Transformer inference on FPGA's and considering the three repositories: IEEE Xplore, Scopus, and arXiv as sources. The scope for this survey is the last two bars (in green).}
	\label{fig:articles_per_year}
\end{figure}

In recent years, growing amounts of research effort has focused on adapting the Transformer model designs to leverage the advantages provided by the FPGA technology. Researchers have explored various optimisation strategies, ranging from quantisation and pruning to custom memory management and parallel processing frameworks, in order to bridge the gap between the theoretical performance of Transformers and the practical limitations posed by FPGA hardware. Furthermore, as plotted in \Cref{fig:articles_per_year}, the number of articles on this topic has risen significantly in the last 2 years, with 153 and 51 articles for 2024 and 2025 till the end of May, respectively. This sharp increase in the number of published articles reflects the increased momentum in channelled research effort.

As such, now can be considered as yet another interesting point in time to observe emerging trends, consider reached conclusions, and identify stated future work approaches. Surveys on the topic have been conducted in the past, with one year frequencies, but not covering the 2024--2025 span. Notably, in works by Kang \textit{et al.}~\cite{Kang:2024:SFAD} and Chitty-Venkata \textit{et al.}~\cite{Chitty-Venkata:2023:STOT}, authors have surveyed the topic extensively. Other, use-case specific surveys exist as well, focusing on the applications on Transformers in specific areas, such as Vision Transformers (ViT)~\cite{Saha:2025:VTEC:arxiv} and Large Language Models (LLMs)~\cite{Koilia:2024:HALC:arxiv}. A detailed coverage of these surveys is listed in \Cref{tab:survey_coverage}.
\begin{table}[htbp]
	\centering
	\caption{Previous surveys and their coverage as a point of comparison to this survey. Note that last listed year, usually the publication year, is never fully covered. These surveys do not specify the exact collection period.}
	\label{tab:survey_coverage}
	\begin{tabular}{c p{1.6cm} lcc}
		\toprule
		\textbf{Year(s)} & \textbf{Platform(s)} & \textbf{Use-case} & \textbf{Systematic} & \textbf{Survey} \\
		\midrule
		--2023 		& ASIC, FPGA,\newline GPU, CPU 	& Non-specific 		& No 	& \cite{Chitty-Venkata:2023:STOT} \\
		--2024 		& ASIC, FPGA 					& Non-specific 		& No 	& \cite{Kang:2024:SFAD} \\
		2025 		& ASIC, FPGA,\newline GPU 		& ViT 				& Yes 	& \cite{Saha:2025:VTEC:arxiv} \\
		--2024 		& ASIC, FPGA,\newline GPU, CPU 	& LLM 				& No 	& \cite{Koilia:2024:HALC:arxiv} \\
		\bottomrule
	\end{tabular}
\end{table}

As the field is extremely active and rapidly evolving, we have opted to focus on the latest developments and achievements. We therefore position this work as a focused recent update rather than a replacement for prior broader surveys. The value of this narrower collection lies in highlighting current implementation tendencies and optimisation choices in a field that is presently moving very quickly. Accordingly, this survey targets all FPGA-based Transformer inference articles published during the year 2024 and the year 2025 till the end of May, using a clear and reproducible selection methodology, i.e., systematic\footnote{Following a predefined, protocol-driven, transparent, and reproducible methodology for study identification, selection, and analysis.}. Our survey does not focus on specific use-case categories.

\subsection{Transformer architecture}
The Transformer architecture, introduced by Vaswani \textit{et al.}~\cite{Vaswani:2017:Attention}, is built entirely on attention mechanisms and is widely used for sequence processing use-cases, e.g., natural language processing and computer vision. At a high level and in its original form, a Transformer consists of an encoder--decoder structure. Many modern applications of the Transformer architecture make use of the encoder block, e.g., BERT, or the decoder block, e.g., GPT-style models, without the other.

Each encoder layer comprises a \emph{multi-head self-attention} segment, followed by a position-wise feed-forward network. The self-attention mechanism allows every token in the input sequence to attend to all other tokens, enabling the model to capture both short- and long-range dependencies. For an input matrix $\mathbf{X}$, three learned linear projections produce the \emph{Query}, \emph{Key}, and \emph{Value} matrices:
\[
\mathbf{Q} = \mathbf{X}\mathbf{W}_Q,\quad
\mathbf{K} = \mathbf{X}\mathbf{W}_K,\quad
\mathbf{V} = \mathbf{X}\mathbf{W}_V,
\]
where $\mathbf{W}_Q$, $\mathbf{W}_K$, and $\mathbf{W}_V$ are the corresponding weight matrices. Accordingly, the attention output is computed as
\[
\mathrm{Attention}(\mathbf{Q}, \mathbf{K}, \mathbf{V}) 
= \mathrm{SoftMax}\!\left(\frac{\mathbf{Q}\mathbf{K}^\top}{\sqrt{d_k}}\right)\mathbf{V},
\]
where $d_k$ is the dimensionality of the key vectors. Through multi-head attention, this computation is replicated across multiple heads, enabling the model to capture diverse relational patterns.

Each decoder layer adds an additional \emph{masked self-attention} segment to the design, to preserve autoregressive behaviour. A cross-attention segment that attends to the encoder outputs is included as well. Both encoder and decoder layers make use of residual connections and layer normalisation~\cite{Ba:2016:LN:arxiv} to ensure stable gradients and improve training efficiency.

\begin{figure}[htbp]
	\centering
	\includegraphics[width=0.6\linewidth]{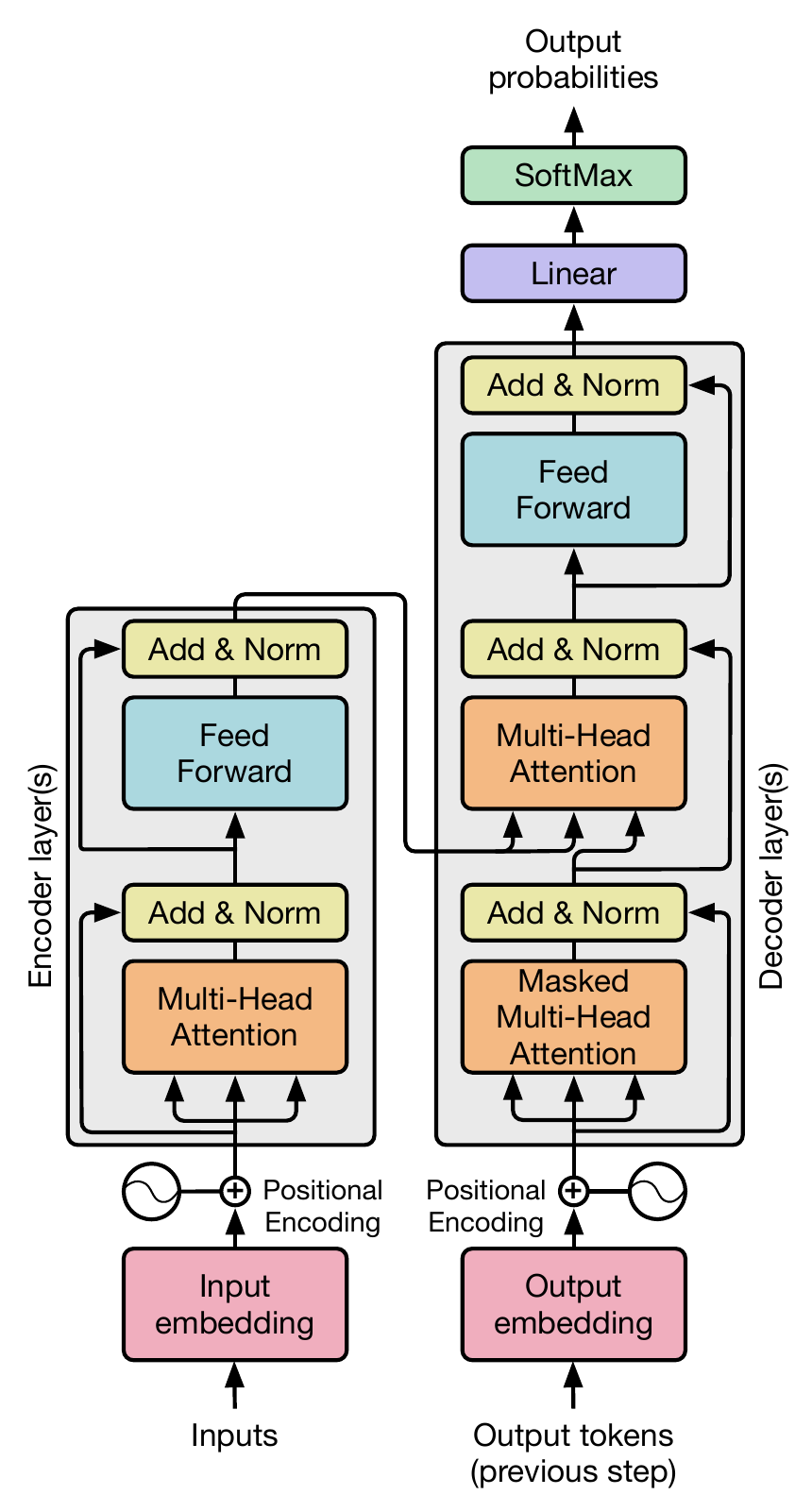}
	\caption{The original Transformer model architecture with encoder and decoder segments.}
	\label{fig:transformer_architecture}
\end{figure}

The attention mechanism is permutation-invariant, for which Transformers rely on \emph{position embeddings} to inject order information into it. These can be either fixed sinusoidal embeddings or learned embeddings. The architectural design eliminates the need for recurrence or convolution, allowing for highly parallel computation. \Cref{fig:transformer_architecture} (inspired by ~\cite{Vaswani:2017:Attention}) depicts the standard Transformer architecture with all the aforementioned segments.

\subsection{Common Transformer model types}
Modern Transformer deployments are typically specialised towards particular data modalities, with Vision Transformers (ViTs) and Large Language Models (LLMs), i.e., generative Transformers, being two dominant categories.

\paragraph*{Vision Transformers} 
ViTs process images by splitting them into fixed-size patches, which are embedded as tokens and fed to a standard Transformer encoder~\cite{Dosovitskiy:2021:IWWT}, i.e., encoder-only. Unlike convolutional models, they rely entirely on global self-attention to model spatial relationships. The main bottleneck for ViTs is the quadratic complexity of self-attention mechanism with respect to the number of patches. While manageable for standard image sizes, this becomes costly for high-resolution inputs. Memory usage is additionally driven by attention maps and intermediate activations across multiple heads and layers.

\paragraph*{Large Language Models} 
LLMs are typically decoder-only, autoregressive models for text generation, e.g., as in GPT-style architectures~\cite{Radford:2019:LMUM}. These models operate on long sequences and contain very large parameter sets. In contrast to ViTs, LLM inference is primarily memory-bound, i.e., model weights are streamed from off-chip memory, making bandwidth a key constraint. Moreover, autoregressive decoding is a limiting factor for parallelism, and the Key--Value (KV) cache introduces additional memory overhead that grows with sequence length.

Overall, ViTs are largely constrained by attention compute complexity, whereas LLMs are dominated by memory capacity, bandwidth, and sequential execution effects. These distinctions directly influence FPGA-oriented optimisation strategies. For this reason, model category is an important dimension when interpreting reported accelerator results and optimisation choices in the later sections of this survey.

\subsection{FPGA technology}
Field-Programmable Gate Arrays (FPGAs) are reconfigurable hardware platforms that allow designers to implement custom digital circuits, tailored to specific computational workloads~\cite{Kuon:2006:MGFA}. Unlike fixed-function Application-Specific Integrated Circuits (ASICs) or general-purpose CPUs, an FPGA consists of an array of programmable logic blocks. Blocks such as Look-Up Tables (LUTs), Flip-Flops (FF), and interconnect resources can be reconfigured after manufacturing. This reconfigurability makes FPGAs attractive to implement purpose-built accelerators, e.g., for accelerating emerging ML architectures, such as Transformers, where computational patterns evolve rapidly~\cite{Venieris:2018:TMCN}.

Modern FPGAs also include a set of hardened resources, providing high-performance arithmetic and memory capabilities. These typically include Digital Signal Processing (DSP) blocks for efficient multiply-accumulate operations, Block RAMs (BRAMs) and UltraRAMs (URAMs) for low-latency on-chip storage, and high-bandwidth interfaces for connecting to off-chip Dynamic Random-Access Memory (DRAM)~\cite{Xilinx:2025:DS890, Intel:2024:Stratix10}. By tailoring computation to these resources, designers can exploit massive fine-grained parallelism and reduce the memory bottlenecks commonly encountered in deep learning workloads~\cite{Zhang:2015:OFAD}.

A key advantage of FPGAs is the ability to define bespoke dataflows and memory hierarchies. Designers can structure computation around streaming architectures, systolic arrays, or pipelined parallel units, depending on the characteristics of the target algorithm~\cite{Boutros:2025:FADL}. On-chip buffers are frequently used to minimise expensive off-chip memory accesses, improving latency. Data reuse strategies, such as tiling, caching, and weight buffering, play a central role in achieving high throughput~\cite{Shen:2017:MCAE}.

FPGA development flows typically rely on Hardware Description Languages (HDLs) such as Verilog or VHDL. However, High-Level Synthesis (HLS) tools have become increasingly common, allowing hardware to be generated from C/C++-like specifications or domain-specific toolchains~\cite{Canis:2011:LegUp, Nane:2016:SEFH}. This has significantly reduced development time and made FPGA-based acceleration more accessible in ML research and industry. This is due to higher level abstractions, i.e., lower implementation complexities, provided by HLS.

Overall, FPGAs offer a flexible and energy-efficient platform for accelerating compute- and memory-intensive tasks. Their customisability, combined with support for high degrees of parallelism, makes them particularly well suited for implementing Transformer components such as attention mechanisms, matrix multiplications, and hierarchical buffering schemes~\cite{Kachris:2025:SHAL}. Needless to say, there is a strong element of use-case specificity involved, making implementation of reusable multi-use designs a constant challenge.

\section{Survey method and pruning protocols}
\label{sec:survey_method}
\begin{figure*}[htbp]
	\centering
	\includegraphics[width=\linewidth]{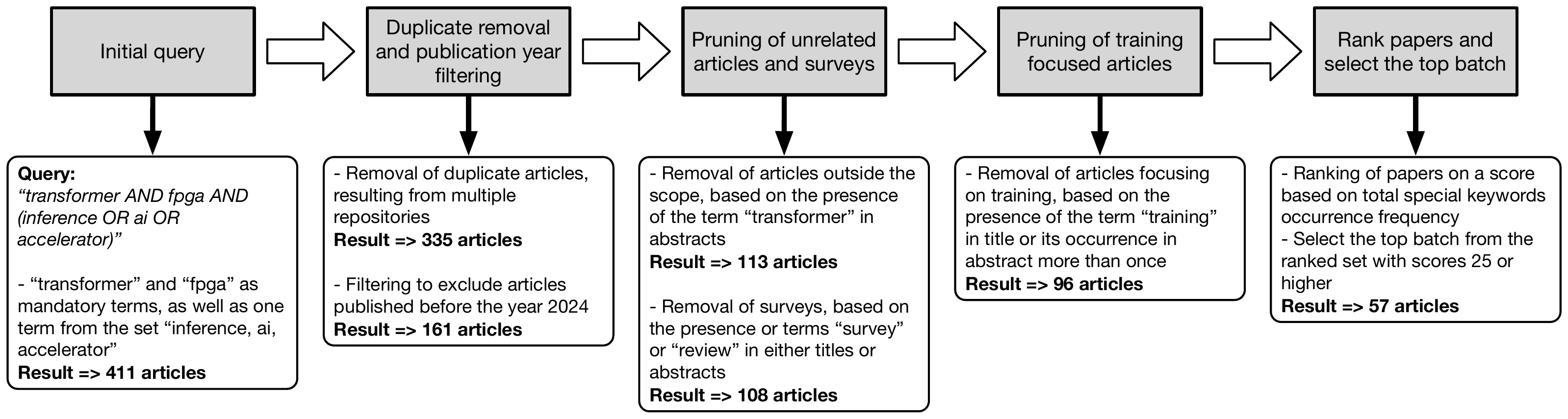}
	\caption{Article collection and pruning workflow, with details on pruning protocols and resulting article counts after each step.}
	\label{fig:article_processing_steps}
\end{figure*}

The literature considered for this review has been gathered and selected in a systematic and semi-automatic fashion. the processing and pruning of collected query responses is performed using Python scripts developed by the authors. The code is available online~\cite{Blankestijn:2026:CODE}. An overview diagram of the steps with article counts resulting from each step is given in \Cref{fig:article_processing_steps}.

\paragraph*{Repositories}
All covered articles are collected from three source repositories, IEEE Xplore\footnote{\url{https://ieeexplore.ieee.org/Xplore/home.jsp}}, Scopus\footnote{\url{https://www.scopus.com/}}, and arXiv\footnote{\url{https://arxiv.org/}}. The collection from arXiv is performed dynamically through the available Application Programming Interface (API)\footnote{\url{https://info.arxiv.org/help/api/index.html}}. These repositories were selected to balance coverage, reproducibility, and accessibility. IEEE Xplore was included for its strong relevance to FPGA and hardware systems research; Scopus was used for its broad cross-publisher indexing coverage (including ACM); and arXiv was included to capture recent preprints in this rapidly evolving field. Together, these sources provide broad visibility of relevant literature while keeping the search protocol transparent and reproducible.

\paragraph*{Query}
A standard query, made up of highly relevant terms was formulated, reflecting our focused topics.
The query,
\begin{quote}
\emph{\enquote{transformer AND fpga AND (inference OR ai OR acceleration)}},
\end{quote}
was used for searching within all three repositories, resulting in a total of 411 articles.

\paragraph*{Pruning - Duplicate removal}
Since multiple repositories are considered and especially in the presence of preprint versions from arXiv, pruning for duplicates is necessary. This pruning step resulted in 335 articles remaining. 

\paragraph*{Pruning - Date filter ($<2024$)}
A date filter applied to bring the focus to publications from 2024 onwards, till the end of May 2025. This pruning step resulted in 161 articles remaining.

\paragraph*{Pruning - Abstract focus}
Subsequently, all articles are filtered based on whether or not the keyword \enquote{transformer} is present in the abstract. This pruning step is to ensure the inclusion of articles with Transformer models as the primary focus. It resulted in 113 articles remaining.

\paragraph*{Pruning - Survey removal}
The response to the query naturally includes surveys relevant to the topic, which are to be set aside. Articles with terms \enquote{survey} or \enquote{review} in either the title or the abstract are considered surveys and filtered. This pruning step resulted in 108 articles remaining.

\paragraph*{Pruning - Articles on training}
Finally, as our aim is to cover inference, articles with a primary focus on training are to be pruned. This is done based on whether
the keyword \enquote{training} is included in the title or if it occurs more than once in the abstract. The remaining articles after this pruning step is 96.

\paragraph*{Ranking}
Collection of articles during any survey will result any a spectrum of articles in terms of relevance. To limit the amount of articles investigated for this review to the ones with the highest relevancy, a ranking is applied to the collection. Points are awarded to each article based on the frequency of keywords, selected from sub-topics relevant to our topic, within the title and the abstract. \Cref{tab:article_ranking_keywords} covers the list of keywords per sub-topic category. The articles are ranked based on the number of points and with 25 as the threshold score. The top 57 articles, achieving 25 points or higher, are considered for this survey. This collection is our primary source. The most relevant of these 57 articles are processed by the authors.
\begin{table}[htbp]
	\centering
	\caption{Keyword groups for ranking articles with scores based on occurrence frequency.}
	\label{tab:article_ranking_keywords}
	\begin{tabular}{l p{0.70\linewidth}}
		\toprule
		\textbf{Category} & \textbf{Keywords} \\
		\midrule
		Synthesis &
		HLS, high level synthesis, high-level synthesis, register transfer level \\
		Memory &
		memory, storage, on-chip, off-chip \\
		Compression &
		compression, quantisation, pruning, sparsity, sparse \\
		Acceleration &
		acceleration, accelerator, reconfigurable devices, systolic, fpga \\
		AI terms &
		ai, artificial intelligence, inference \\
		Misc.\ terms &
		weight, weights, computation, efficient, constrain \\
		\bottomrule
	\end{tabular}
\end{table}

\section{Taxonomy of Transformer inference on FPGAs}
\label{sec:taxonomy}
\begin{figure}[htbp]
	\centering
	\includegraphics[width=0.9\linewidth]{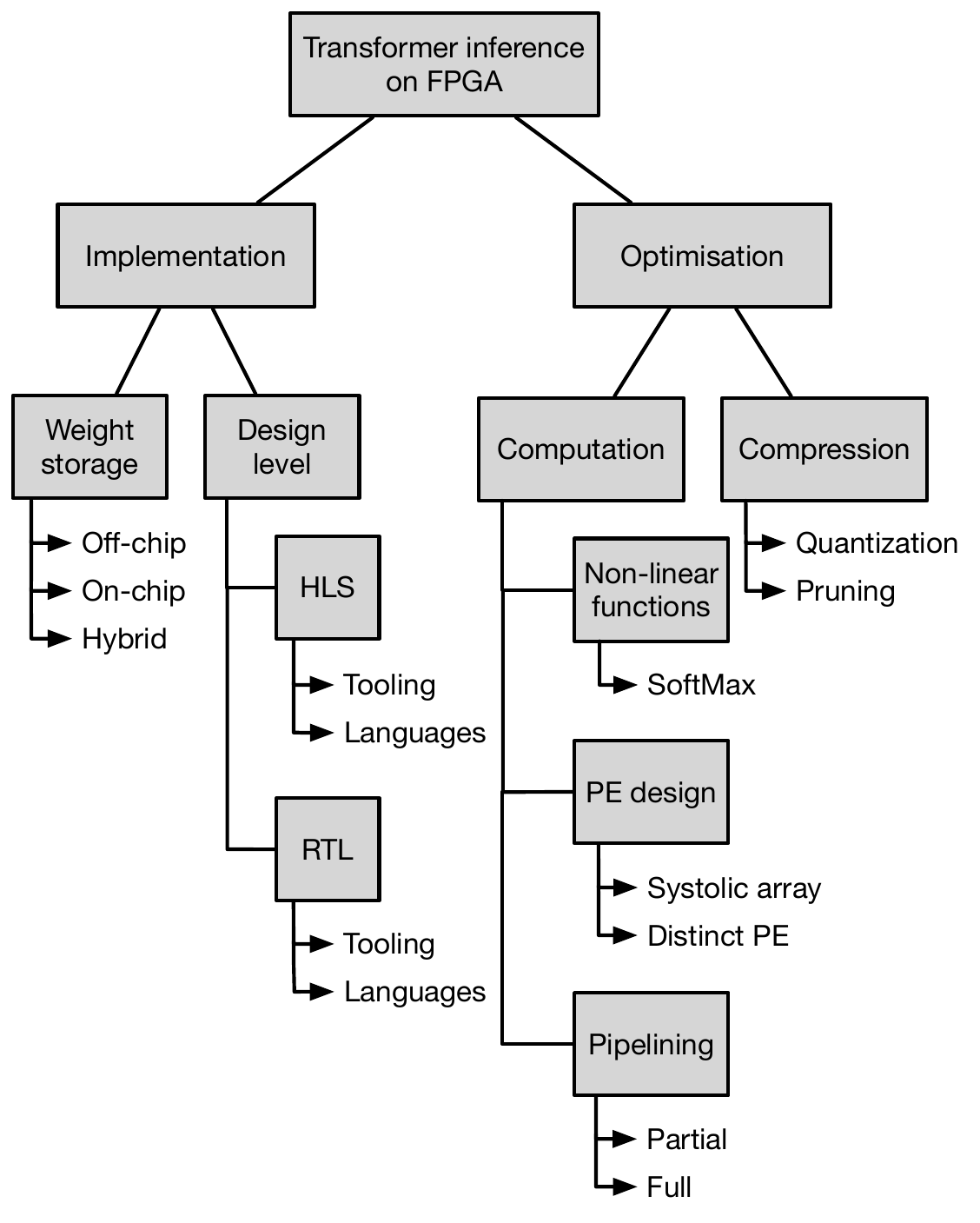}
	\caption{Extracted taxonomy of prevalent approaches for Transformer inference on FPGA platforms, summarising the main subjects, design dimensions, and optimisation techniques identified in the surveyed literature.}
	\label{fig:topics_taxonomy}
\end{figure}

Given the collected articles and based on our review of these, we provide a taxonomy of approaches and techniques covered by these literature while deploying Transformer models on FGPA platforms for inference, drawn in \Cref{fig:topics_taxonomy}.

The taxonomy provides a natural hierarchy, encompassing the two main aspects addressed in the literature, \emph{Implementation} and \emph{Optimisation}. We shall use this hierarchy to organise the remainder of this survey. The taxonomy is derived from the analysis of recurring themes and design choices from the collected articles. The authors have incorporated their knowledge of similar implementations to arrive at a more meaningful structure. While not exhaustive, \Cref{fig:topics_taxonomy} reflects the dominant implementation and optimisation techniques reported and serves as a conceptual guide for positioning individual contributions. The performed ranking of articles based on relevancy to the topic elevates the taxonomy higher in its representativeness.

\section{Model weights storage}
\label{sec:model_weights_storage}
Memory transfer can be a big bottleneck in inference latency. Therefore, the way that the model weights and biases are stored and handled can have a big impact on performance. Various techniques exist for storing the weights and biases of a model. The two most popular approaches are storing all weights on-chip or dynamically loading parameters from off-chip memory, as needed.

Ideally, all parameters are permanently stored on-chip, i.e., on the FPGA, in order to minimise memory transfer. However, this is not always feasible when dealing with large models and/or smaller FPGAs. Additionally, this approach can be impractical for generic accelerators aimed at supporting multiple types of models. As such, a rather common technique is to store the weights in external memory or have the weights loaded dynamically trough the host CPU. Note that most hardware platforms provide a CPU, often time an ARM core, alongside the FPGA, hence the host CPU.

\subsection{Off-chip storage}
Accelerators making use of off-chip memory access are generally capable of supporting many different models and model sizes, since these are not limited by the available on-chip memory. In one such approach, Kabir \textit{et al.}~\cite{Kabir:2024:RATN:arxiv} propose a Transformer accelerator that is runtime-adaptive and makes use of off-chip memory access. The architecture makes use of separate \emph{Load Units} for inputs, weights, and biases. For instance, on-chip weight buffers for the attention computation are 2-dimensional arrays, where the array size depend on the tiling size. For each iteration when handling each tile, the partial data is loaded from external memory. Since biases are generally smaller in size, these are stored in registers. The biases are loaded into registers while the attention module is computing the $\mathbf{Q}$, $\mathbf{K}$, and $\mathbf{V}$ matrices. Afterwards, the biases are then added to the matrices. FAMOUS~\cite{Kabir:2024:FAMOUS:arxiv} is another accelerator from the same authors that employs a very similar system. The weight matrices are loaded in on-chip memory for every iteration and the biases are transferred to internal registers during the matrix multiplication step.

Optimisations to reduce the number of memory transfers can still be made, even when making use of off-chip memory. For example, Li and Chen~\cite{Li:2025:DIFH:arxiv} propose a \emph{Tiled Matrix Multiplication Accelerator} for self-attention in Transformers. This accelerator focuses only on the $\mathbf{Q}$, $\mathbf{K}$, and $\mathbf{V}$ projections. For each matrix multiplication, the $\mathbf{Q}$, $\mathbf{K}$, and $\mathbf{V}$ weight matrices are transferred to the FPGA. However, in order to minimise memory transfers, the input matrix is only transferred to the FPGA once for the $\mathbf{Q}$ projection. Then, the input matrix is stored in an on-chip buffer for $\mathbf{K}$ and $\mathbf{V}$ projections.

\subsection{On-chip storage}
The most performance-effective way to store a model's parameters is to store these on-chip. On-chip storage minimises off-chip memory access, which in turn has a big impact on overall performance. As mentioned, considering limited on-chip memory, storing all weights and activations on-chip is not always possible. Another advantage of permanently storing weights using on-chip storage is to facilitate the use of extensive pipelining, as exemplified by Guo \textit{et al.}\ with HG-PIPE~\cite{Guo:2024:HG-PIPE:arxiv}. In order to take advantage of pipelining, the authors of HG-PIPE propose an implementation that stores as many weights and activations on-chip as possible.

Alternatively, instead of having all weights fixed in on-chip memory, it is also possible to store these in on-chip buffers with the possibility of loading in different weights for different models. Take ME-ViT for instance, authored by Marino~\cite{Marino:2024:ME-ViT:arxiv}, which is an accelerator for Vision Transformers. This accelerator will load in all parameters from off-chip memory and store these in on-chip buffers before starting the inference. Parameters that have to be re-used multiple times during a single inference are strategically buffered in on-chip memory. Then, the Memory Efficient Processing Element (ME-PE) is able to perform in 3 different modes: Linear Projection, Multi-Headed Self-Attention and Multi-Layer Perceptron. Any intermediate results during inference are also kept in on-chip buffers. This architecture ensures that there are minimal amount of memory transfers between off-chip and on-chip variants. This implementation does suffer from on-chip memory limitations. It must be noted that this architecture does allow for inference of multiple different models, compared to storing of parameters permanently on-chip.

Yet another example of an accelerator storing all the model weights and activations in on-chip memory is ProTEA, authored by Kabir \textit{et al.}~\cite{Kabir:2024:ProTEA}. The authors propose an efficient tiling strategy, allowing large models to be stored in on-chip memory. ProTEA is runtime configurable, enabling inference of different models. All the model weights are loaded into on-chip memory ahead of the inference.

\subsection{Performance comparison}
\Cref{tab:weights_storage_approach_comparison} presents a comparison in resource consumption between articles employing off-chip memory access and articles storing model weights on-chip during inference. It is important to note that a considerable number of articles do not make it clear if the incorporated accelerator loads in all weights from off-chip memory into on-chip memory prior to inference, or if the implementation accesses off-chip memory during inference. We would qualify the former approach as \enquote{on-chip memory storage} as there would be no off-chip memory access during the inference itself. Therefore, it is likely that a significant portion of implementations that we have categorised as \enquote{off-chip} are actually qualify as \enquote{on-chip}. It is difficult to make precise comparisons here. Nevertheless, it is interesting to see if a trend can be noticed.

As it can be seen in \Cref{tab:weights_storage_approach_comparison}, accelerators making use of off-chip memory access generally have a higher resource consumption. With significantly more FF, LUTs, DSPs and BRAM usage. At the same time, accelerators keeping all the weights in on-chip memory have a significantly higher throughput and much lower latency. This may seem illogical, as one would expect that implementations with higher resource consumption should result in better performance. However, this can be explained by the memory access time required due to the incorporation of off-chip memory, which is a considerable bottleneck, severely limiting both throughput and latency.
\begin{table*}[htbp]
	\centering
	\caption{Resource utilisation and performance comparison between storage approaches, for off-chip and on-chip. 6 articles with unknown storage approach are not covered here. Note that not every field is reported in every article. Cells designated with \enquote{n/a} indicate data that were unavailable or not reported in the original paper.}
	\label{tab:weights_storage_approach_comparison}
	\begin{tabular}{cccc|l|ccc|c}
        \toprule
        \textbf{FF} & 
        \textbf{LUT} & 
        \textbf{DSP} & 
        \textbf{BRAM} & 
        \textbf{Platform} &
        \textbf{Throughput (GOPS)} & 
        \textbf{Power (W)} & 
        \textbf{Latency (ms)} &
        \textbf{Article} \\
        \midrule
		\multicolumn{9}{c}{\textbf{Storage approach: Off-chip}} \\
		\midrule
		114K 			& 128K 			& 768 		& n/a 			& ZCU102 		& 1\,023.3 		& 23.48 	& n/a 		& \cite{Zhao:2025:HEAT} \\
		n/a 			& 77K 			& 147 		& n/a 			& VCK190 		& n/a 			& n/a 		& n/a 		& \cite{Zhong:2024:FYET} \\
		n/a 			& n/a 			& 1\,024 	& n/a 			& ZCU102 		& 780 			& 7.43 		& n/a 		& \cite{Shao:2024:FRAC} \\
		1\,507K 		& 557K 			& 5\,096 	& 1\,581 		& XCU200 		& 2\,750 		& 28.23 	& 0.08 		& \cite{Zhang:2024:FNM-Trans} \\
		n/a 			& 271K 			& 1\,863 	& 609.5 		& Alveo U50 	& 301.9 		& 14.35 	& n/a 		& \cite{Liu:2023:EFAS:arxiv} \\
		n/a 			& 391K 			& 3\,612 	& n/a 			& n/a 			& 40 			& 11.9 		& n/a 		& \cite{Kabir:2024:RATN:arxiv} \\
		943K 			& 574K 			& 6\,345 	& 1\,253 		& Alveo U280 	& n/a 			& n/a 		& n/a 		& \cite{Zeng:2024:FlightLLM} \\
		n/a 			& n/a 			& n/a 		& n/a 			& Alveo U55C 	& n/a 			& n/a 		& n/a 		& \cite{Bai:2024:SWAT:arxiv} \\
		n/a 			& n/a 			& n/a 		& n/a 			& n/a 			& n/a 			& 9 		& 17.51 	& \cite{He:2024:HLSTransform:arxiv} \\
		704K 			& 993K 			& 3\,612 	& n/a 			& VCK190		& 44 			& n/a 		& 165 		& \cite{Kabir:2024:ProTEA} \\
		222K 			& 115K 			& 1\,248 	& n/a 			& KV260			& n/a 			& 10 		& 50 		& \cite{Li:2024:MPTA} \\
		143K 			& 123K 			& 1\,850 	& 458 			& ZCU102		& 97.04 		& 11.5 		& 25.76 	& \cite{Dong:2025:UbiMoE} \\
		2\,446K 		& 1\,593K 		& 10\,848 	& 1\,746 		& AMD V80		& n/a 			& n/a 		& n/a 		& \cite{Liu:2025:FlightVGM} \\
		n/a 			& n/a 			& n/a 		& n/a 			& ZCU102		& 385.5 		& 9.86 		& n/a 		& \cite{Chen:2024:SALTS} \\
		162K 			& 122K 			& 78 		& 691.5 		& ZCU102		& 3\,894.74 	& 8.68 		& n/a 		& \cite{Qiao:2025:COBRA:arxiv} \\
		n/a 			& n/a 			& n/a 		& n/a 			& VCK190		& n/a 			& 47.5 		& 56 		& \cite{Dong:2024:EQ-ViT} \\
		n/a 			& 493K 			& 5\,016 	& n/a 			& Alveo U280	& 1\,223.9 		& n/a 		& 14.57 	& \cite{Wang:2024:TransFRU} \\
		n/a 			& n/a 			& 3\,482 	& 1\,162 		& Alveo U50		& 1\,420 		& 48 		& n/a 		& \cite{Qin:2024:ELSI} \\
		578K 			& 472K 			& 9\,024 	& 1\,520 		& VCU128		& 8\,204 		& 43.2 		& n/a 		& \cite{He:2025:F3} \\
		661K 			& 1\,284K 		& 4\,157 	& 3\,148 		& Alveo U55C	& 184 			& n/a 		& 0.597 	& \cite{Kabir:2024:FAMOUS:arxiv} \\
		103K 			& 71K 			& 1\,050 	& 126 			& KV260			& 2.85 			& n/a 		& 110 		& \cite{Li:2025:DIFH:arxiv} \\
		\hline
		682\,500 		& 506\,055.56 	& 3\,842 	& 1\,200.15 	& --			& 1\,453.66 	& 20.075 	& 49.9177 	& Total average \\
		\midrule
		\multicolumn{9}{c}{\textbf{Storage approach: On-chip}} \\
		\midrule
		n/a 			& n/a 			& n/a 		& n/a 			& Alveo U200 	& n/a 			& 31.8 		& n/a 		& \cite{Marino:2024:ME-ViT:arxiv} \\
		73K 			& 110K 			& 0 		& 177 			& ZCU102		& 944.87 		& 1.39 		& n/a 		& \cite{Du:2024:EEFB} \\
		n/a 			& 669K 			& 312 		& 1\,006 		& VCK190		& 17\,795 		& 46.7 		& n/a 		& \cite{Guo:2024:HG-PIPE:arxiv} \\
		291K 			& 161K 			& 1945 		& 812 			& ZCU102		& n/a 			& 16.71 	& 1.77 		& \cite{Byun:2024:HFHR} \\
		548K 			& 274K 			& 2520 		& 912 			& ZCU102		& 1\,387.59 	& 7.95 		& n/a 		& \cite{Ji:2024:BETA} \\
		139K 			& 118K 			& 2147 		& 283 			& XCZU9EG		& 2\,330.2 		& 21.37 	& 13.98 	& \cite{Zhang:2024:FVTA} \\
		1\,083K 		& 544K 			& 10\,812 	& 1\,903 		& Alveo U250 	& 986.3 		& n/a 		& 1.17 		& \cite{Sun:2025:DRViT} \\
        \hline
        426\,800 		& 312\,666.67 	& 2\,956 	& 848.83  		& --			& 4\,688.79 	& 20.987 	& 5.64 		& Total average \\
        \bottomrule
    \end{tabular}
\end{table*}

We can conclude from \Cref{tab:weights_storage_approach_comparison} that in general, utilising as much persistent on-chip storage as possible results in much lower inference latencies and higher throughput. However, if the goal is to make an accelerator suitable for a wider variety of models, then the only possible architecture could be one where the weights are not stored on-chip, but rather off-chip.

It should be noted that \Cref{tab:weights_storage_approach_comparison} focuses specifically on the relation between storage approach, resource usage, and performance. The primary optimisation techniques employed by the corresponding accelerators, such as quantisation, sparsity, and architectural restructuring, are discussed separately in the later comparative analysis, particularly in \Cref{tab:top_ten_troughput,tab:top_ten_efficiency}.

\section{Incorporated tooling}
\label{sec:incorporated_tooling}
Various techniques are available to develop solutions for FPGAs. These techniques might vary depending on the model or manufacturer of the considered FPGA. We can categorise the available tooling and programming languages under two main existing design levels, Register-Transfer Level (RTL) and High-Level Synthesis (HLS). HLS tools simplify development substantially, as there is no requirement for knowledge of low-level RTL-design and hardware description languages. The opposite is the case with RTL tooling, with a focus on low-level hardware design, which also provides the opportunity for ad hoc optimisations. Considering these two design level, i.e., design methodologies, the prevalence of tooling from each is investigated below, alongside the performance and resource consumption impact analysis of selected tooling.

\Cref{tab:fpga_tooling} lists how often each type of tooling is used within the analysed pool of articles in this survey. The table is divided into two categories: RTL and HLS. The RTL category includes Verilog, SystemVerilog, and VHDL, while the HLS category includes Vitis, Versel ACAP, hls4ml, and Unknown tooling. The Unknown category is for articles not specifying the utilised tooling.
\begin{table}[htbp]
	\centering
	\caption{Comparison of synthesis tooling used in different article, with a significant number lacking elaboration.}
	\label{tab:fpga_tooling}
	\begin{tabular}{p{1.5cm} lc p{2.5cm}}
		\toprule
		\textbf{Design\newline level} & \textbf{Language/Tool} & \textbf{Count} & \textbf{Articles} \\
		\midrule
		\multirow{3}{*}{RTL}
									& Verilog        		& 10 	& \cite{Wang:2024:EFTA, Li:2024:EEFA, Zhao:2025:HEAT, Shao:2024:FRAC, Du:2024:EEFB, Byun:2024:HFHR, Chen:2024:SALTS, Wang:2024:TransFRU, He:2025:F3, Zhang:2024:VisionAGILE} \\
									& SystemVerilog 		& 1 	& \cite{Li:2024:MPTA} \\
									& VHDL           		& 1 	& \cite{Ling:2024:IQTE} \\
		\hline
		\multirow{4}{*}{HLS}
									& Vivado/Vitis HLS 		& 11 	& \cite{Marino:2024:ME-ViT:arxiv, Zhong:2024:FYET, Kabir:2024:RATN:arxiv, Bai:2024:SWAT:arxiv, He:2024:HLSTransform:arxiv, Kabir:2024:ProTEA, Dong:2025:UbiMoE, Qiao:2025:COBRA:arxiv, Qin:2024:ELSI, Kabir:2024:FAMOUS:arxiv, Li:2025:DIFH:arxiv} \\
									& Versal ACAP    		& 1 	& \cite{Dong:2024:EQ-ViT} \\
									& hls4ml         		& 1 	& \cite{Jiang:2024:LLTI:arxiv} \\
									& Unknown        		& 2 	& \cite{Guo:2024:HG-PIPE:arxiv, Sadeghi:2024:CHOSEN:arxiv} \\
		\hline
		Unknown 					& Unknown 				& 11 	& \cite{Xiang:2025:ELIM:arxiv, Zhang:2024:FNM-Trans, Zeng:2024:FlightLLM, Ji:2024:BETA, Liu:2025:FlightVGM, Zhang:2024:FVTA, Sun:2025:DRViT, Dong:2024:SWAT, Seo:2024:IANUS, Xu:2024:FQAV, Moitra:2024:PIVOT} \\
		\bottomrule
	\end{tabular}
\end{table}

Unfortunately, not all articles specify the incorporated tooling. For the articles that do specify it, the distribution between the usage of RTL tooling and HLS tooling is rather evenly split, with 12 articles making use of RTL tooling and 13 utilising HLS tooling. Within RTL tooling, the Hardware Description Language (HDL) Verilog\footnote{Also standardised as IEEE 1364: \url{https://standards.ieee.org/ieee/1364/2052/}} is by far the most popular choice. Only 2 articles used other tooling, namely~\cite{Li:2024:MPTA} and~\cite{Ling:2024:IQTE} using SystemVerilog and VHDL, respectively. A similar trend can be seen within HLS tooling, where Vitis HLS is by far the most popular choice. Other tooling used includes Versel ACAP (by~\cite{Dong:2024:EQ-ViT}) and hls4ml (by~\cite{Jiang:2024:LLTI:arxiv}). 

\subsection{Frameworks}
A small selection of articles set out to develop or extended frameworks for the development of Transformer accelerators. 
Jiang \textit{et al.}~\cite{Jiang:2024:LLTI:arxiv} extended the popular HLS tool, hls4ml~\cite{Duarte:2018:FIDN, FastML:2025:hls4ml} to include support for any Tensorflow-based Transformer model. hls4ml is an open-source Python package that can translate machine learning models from frameworks like Keras and PyTorch into HLS code for deployment on FPGAs. The authors of~\cite{Jiang:2024:LLTI:arxiv} have used Vivado in the implementation of the Transformer for hls4ml. 

On the RTL side, Ling \textit{et al.}~\cite{Ling:2024:IQTE} developed VHDL templates which enable developers to translate existing models onto an FPGA accelerator, without a deep understanding of FPGA development workflow. In order to generate an accelerator, Python scripts translate these models using VHDL templates, resulting in VHDL files. This approach ensures a seamless translation of trained and quantised models into FPGA accelerators. The authors note that in the future they aim to support mixed-precision quantisation and to improve the energy efficiency of resulting accelerators.

\subsection{HLS vs RTL}
\Cref{tab:rtl_hls_tooling_comparison} provides a comparison of the average resource usage, as well as performance metrics, i.e., throughput in Giga Operations Per Second (GOPS) and power in Watts. The comparison is between articles using RTL design level tooling with the ones utilising HLS design level tooling. It is important to consider that the architecture and design of the accelerator at hand has a much bigger impact on these figures. However, these figures can portray an overall trend. For instance, articles making use of HLS tooling, on average seem to have much more FF and LUT consumption. Additionally, the power consumption is on average twice that of articles making use of RTL tooling. 

While this suggests that RTL-based designs are generally more efficient, the underlying cause is not entirely clear. This trend may either stem from the greater level of control offered by RTL-level design, or from the fact that low-power use-cases are more likely to favour RTL implementations. Regardless of the underlying reason, the observed pattern remains the same, i.e., articles incorporating RTL tooling tend to occupy the lower end of the power spectrum. Moreover, as a lower-level design approach, RTL tooling inherently provide greater potential for granular optimisation.

Interestingly, the average throughput is rather higher for HLS tooling designs. One possible explanation is that HLS tooling allows more time and effort to be spent on efficient computations, i.e., enabling the tooling to convert the model design into the best possible RTL design. Similarly, \Cref{tab:rtl_hls_tooling_comparison} is intended to compare tooling-related trends rather than the underlying optimisation techniques used in each accelerator. Design-specific innovations, including quantisation strategies, sparsity, memory restructuring, and kernel-level optimisations, are treated separately in the subsequent performance-oriented comparison.

\begin{table*}[htbp]
	\centering
	\caption{Average resource usage and performance metrics, covering throughput (in GOPS) and power (in Watts), compared between RTL vs HLS design level tooling. Note that not every field is reported in every article. Cells designated with \enquote{n/a} indicate data that were unavailable or not reported in the original paper.}
	\label{tab:rtl_hls_tooling_comparison}
	\begin{tabular}{cccc|l|cc|c}
		\toprule
		\textbf{FF} &
		\textbf{LUT} &
		\textbf{DSP} &
		\textbf{BRAM} &
		\textbf{Platform} &
		\textbf{Throughput (GOPS)} &
		\textbf{Power (W)} &
		\textbf{Article} \\
		\midrule
		\multicolumn{8}{c}{\textbf{Tooling category: RTL}} \\
		\midrule
		169K 			& 125K 			& 1\,032 		& n/a 		& ZCU102		& 695.37 		& 4.78 		& \cite{Wang:2024:EFTA} \\
		152K 			& 208K 			& 832 			& n/a 		& ZCU102		& 634.27 		& 3.72 		& \cite{Li:2024:EEFA} \\
		114K 			& 128K 			& 768 			& n/a 		& ZCU102		& 1\,023.3 		& 23.48 	& \cite{Zhao:2025:HEAT} \\
		n/a 			& n/a 			& 1\,024 		& n/a 		& ZCU102		& 780 			& 7.43 		& \cite{Shao:2024:FRAC} \\
		73K 			& 110K 			& 0 			& 177 		& ZCU102		& 944.87 		& 1.39 		& \cite{Du:2024:EEFB} \\
		291K 			& 161K 			& 1\,945 		& 812 		& ZCU102		& n/a 			& 16.71 	& \cite{Byun:2024:HFHR} \\
		222K 			& 115K 			& 1\,248 		& n/a 		& KV260			& n/a 			& 10 		& \cite{Li:2024:MPTA} \\
		n/a 			& 1\,008K 		& 9\,091 		& 1\,819 	& Alveo U250	& n/a 			& n/a 		& \cite{Nechi:2023:FDLI} \\
		n/a 			& n/a 			& n/a 			& n/a 		& ZCU102		& 385.5 		& 9.86 		& \cite{Chen:2024:SALTS} \\
		n/a 			& 493K 			& 5\,016 		& n/a 		& Alveo U280	& 1\,223.9 		& n/a 		& \cite{Wang:2024:TransFRU} \\
		578K 			& 472K 			& 9\,024 		& 1\,520 	& VCU128		& 8\,204 		& 43.2 		& \cite{He:2025:F3} \\
		n/a 			& n/a 			& n/a 			& n/a 		& XC7S15		& n/a 			& 0.075 	& \cite{Ling:2024:IQTE} \\
		\hline
		228\,428 		& 313\,555 		& 2\,998 		& 1\,082 	& --			& 2\,093 		& 13 		& Total average \\
		\midrule
		\multicolumn{8}{c}{\textbf{Tooling category: HLS}} \\
		\midrule		
		n/a 			& n/a 			& n/a 			& n/a 		& Alveo U200	& n/a 			& 31.8 		& \cite{Marino:2024:ME-ViT:arxiv} \\
		n/a 			& 77K 			& 147 			& n/a 		& VCK190		& n/a 			& n/a 		& \cite{Zhong:2024:FYET} \\
		n/a 			& 669K 			& 312 			& 1\,006 	& VCK190		& 17\,795 		& 46.7 		& \cite{Guo:2024:HG-PIPE:arxiv} \\
		n/a 			& 391K 			& 3\,612 		& n/a 		& n/a			& 40 			& 11.9 		& \cite{Kabir:2024:RATN:arxiv} \\
		n/a 			& n/a 			& n/a 			& n/a 		& Alveo U55C	& n/a 			& n/a 		& \cite{Bai:2024:SWAT:arxiv} \\
		n/a 			& n/a 			& n/a 			& n/a 		& n/a			& n/a 			& 9 		& \cite{He:2024:HLSTransform:arxiv} \\
		704K 			& 993K 			& 3\,612 		& n/a 		& Alveo U55C	& 44 			& n/a 		& \cite{Kabir:2024:ProTEA} \\
		143K 			& 123K 			& 1\,850 		& 458 		& ZCU102		& 97.04 		& 11.5 		& \cite{Dong:2025:UbiMoE} \\
		162K 			& 122K 			& 78 			& 691.5 	& ZCU102		& 3\,894.74 	& 8.68 		& \cite{Qiao:2025:COBRA:arxiv} \\
		n/a 			& n/a 			& n/a 			& n/a 		& VCK190		& n/a 			& 47.5 		& \cite{Dong:2024:EQ-ViT} \\
		n/a 			& n/a 			& 3\,482 		& 1\,162 	& Alveo U50		& 1\,420 		& 48 		& \cite{Qin:2024:ELSI} \\
		607K 			& 833K 			& 6\,225 		& 1\,440 	& VU9P			& n/a 			& n/a 		& \cite{Han:2023:HPTA} \\
		661K 			& 1\,284K 		& 4\,157 		& 3\,148 	& Alveo U55C	& 184 			& n/a 		& \cite{Kabir:2024:FAMOUS:arxiv} \\
		n/a 			& n/a 			& n/a 			& n/a 		& n/a			& n/a 			& n/a 		& \cite{Jiang:2024:LLTI:arxiv} \\
		103K 			& 71K 			& 1\,050 		& 126 		& KV260			& 2.85 			& n/a 		& \cite{Li:2025:DIFH:arxiv} \\
		\hline
		396\,666 		& 507\,000 		& 2\,452 		& 1\,147 	& --			& 2\,934 		& 26.885 	& Total average \\
		\bottomrule
	\end{tabular}
\end{table*}

\section{Model compression}
\label{sec:model_compression}
Various model compression techniques exist in order to reduce model size, enabling efficient usage of on-chip storage instead of off-chip storage. Furthermore, certain pruning techniques can reduce the model size and complexity, which in turn can result in increased inference performance. Popular methods include \emph{quantisation} and \emph{pruning}. The following covers and compares commonly used quantisation and pruning techniques. This includes comparing different bit-widths and if the same quantisation levels can be used across different types of parameters and layers. The impact on latency, throughput and accuracy of compression techniques must be taken into account as well.

\subsection{Quantisation}
Using 8-bit quantisation is by far the most common practice in articles using quantisation, 13 out of 30. Some papers use higher bit-widths, e.g., 12 or 16 bits (\cite{Dong:2024:SWAT} and \cite{Zhang:2024:FNM-Trans, Qin:2024:ELSI}, respectively), in order to retain a higher accuracy. On the contrary, other implementations opt for even lower bit-widths in order to optimise and reduce computational latency even further. Another approach is to make use of mixed quantisation, e.g.,~\cite{Zeng:2024:FlightLLM, Li:2024:MPTA}. In this approach, important parameters are quantised with a higher bit-width, whereas parameters that have less influence on the final accuracy are quantised with a lower bit-width.

HG-PIPE, by Guo \textit{et al.}~\cite{Guo:2024:HG-PIPE:arxiv}, makes extensive use of LUT-based MAC units, i.e., multiply–accumulate operators implemented using FPGA lookup tables instead of dedicated DSP blocks. If the operands are quantised to 3 bits, the multiplication operation can be decomposed into six boolean functions, each consuming 6 bits. Only 6 LUT-6s are required for such an operation. This technique reduces the number of required DSPs from more than 10\,000 to only 744. Furthermore, reducing the bit-width from 4 to 3 only reduces the final accuracy from 73.30\% to 69.60\%.

Jiang \textit{et al.}\ in~\cite{Jiang:2024:LLTI:arxiv} have extended the hls4ml framework~\cite{FastML:2025:hls4ml} to support the Transformer model. hls4ml allows for varying precision across different layers. The authors have tested various fixed precision configurations, including both Post Training Quantisation (PTQ) and Quantisation Aware Training (QAT) across three different Transformer models. The authors settled on using 6 bits for two of these models and 10 bits for the last model. Using any lower bit-widths would result in a significant drop in the Area Under the Receiver Operating Characteristics curve (AUC). This shows that the optimum quantisation bit-width can heavily depend on the model under evaluation. 

Du \textit{et al.}~\cite{Du:2024:EEFB} propose an accelerator for binary Transformers. In this case, all model parameters are 1-bit quantised. To be precise, all weighs are binarised to \{-1, 1\}, activations of ReLU and SoftMax are binarised to \{0, 1\}, and all other activations are binarised to \{-1, 1\}. The resulting architecture uses significantly fewer resources and provides higher throughput compared to other sparsity-aware accelerators. The major downside to this approach is that it is not feasible to quantise a model to 1-bit during post-training quantisation.

Xiang \textit{et al.}~\cite{Xiang:2025:ELIM:arxiv} make use of an Activation-aware Weight Quantisation (AWQ) technique, in order to significantly reduce the memory footprint. This technique is based on the fact that only 1\% of the model weights may have a significant impact on model performance. Thus, preserving these weights can maintain accuracy while reducing the memory footprint. The technique works by performing scaling on a per-channel basis, based on the activation distribution. This way, the entire matrix can be quantised to very lower precisions (INT4, INT3) while maintaining model accuracy. Evaluation of this quantisation technique results in a 55.10\% reduction in memory footprint with only a 2.82 percentage points reduction in model accuracy compared to a baseline. 

HEAT by Zhao \textit{et al.}~\cite{Zhao:2025:HEAT} is a Transformer Accelerator employing a hybrid-precision quantisation scheme. The resulting model weights are quantised to an average of 5.71 bits, with only a 0.526\% accuracy loss for the model. This technique works by splitting the weight matrices based on a predefined threshold. Values within this threshold are considered \enquote{normal} and values outside of the threshold are considered outliers. The normal values are quantised to 5 bits while the outliers are quantised to 8 bits. Choosing a suitable threshold value is important. In prior work the threshold is set to a constant value. However, in~\cite{Zhao:2025:HEAT}, the threshold is determined by a line search based on a constant outlier ratio of 3.5\%.

Byun \textit{et al.}\ use a Hessian-driven quantisation approach~\cite{Dong:2019:HAWQ} in~\cite{Byun:2024:HFHR}. The authors note that using such an approach has challenges, such as the fact that quantised weights using a parameter-wise approach result in irregular precision patterns. This is not ideal for hardware accelerators. To overcome such limitations, the authors propose a row-wise approach. This technique works by splitting the matrices into important and unimportant rows. The aggregate parameter sensitivity is determined for each row within every matrix of each layer. The rows are sorted based on their sensitivity. Then, all weights are quantised with lower precision while quantising only the top 1\% of weight rows with a higher precision. This process is repeated iteratively until the desired level of accuracy is achieved. The architecture of the accelerator is designed to be able to efficiently handle both the low- and the high-precision calculations. The resulting accelerator increases energy efficiency up to 14.57 times compared to existing accelerators. 

Another method for mixed precision quantisation is implemented by Li \textit{et al.}~\cite{Li:2024:MPTA}. This article implements a Transformer model for the task of license plate recognition. In order to minimise memory overhead, the authors propose a mixed precision quantisation method, optimising the allocation of low bit-width for each layer. Additionally, a Dynamic Precision Adaption (DPA) algorithm is proposed to optimise the allocation between the integer and fractional part. This algorithm considers the magnitude of input and allocates the appropriate bit-width for the largest input value in order to avoid overflow. The resulting algorithm increases the model weight compression by 2.44 times, as opposed to 1.95 times compression when using 8-bit fixed point quantisation, while maintaining identical model accuracy.

\subsection{Model pruning}
Li \textit{et al.}~\cite{Li:2024:EEFA} note that the output of the inner product of the $\mathbf{Q}$ and $\mathbf{K}$ matrices can contain up to 95\% zeroes or close to zero values. These terms have little contribution to the overall model accuracy. Eliminating the computation of these products can significantly reduce the amount of computation, while not affecting the model accuracy by much. The sparsity cannot be predicted before the actual matrix computation though. However, the authors propose a dynamic sparse computation method. First, a low-precision 4-bit computation between the $\mathbf{Q}$ and $\mathbf{K}$ matrices is performed. This results in a mask matrix, indicating positions of non-zero values. This mask is used to select the $\mathbf{K}$ matrix during full 16-bit computation. Compared to other accelerators, this implementation can result in improved throughput and energy efficiency. 

Wang~\cite{Wang:2024:EFTA} proposes a block-wise balanced pruning technique for model compression. The technique works by first splitting up the parameter matrix into smaller blocks of predetermined sizes. Then, pruning is performed for each block, such that each block has the same number of parameters removed. Pruning is done by ranking parameters in descending order. The smallest P parameters are removed depending on the pruning ratio, P. The second step is to use an efficient memory storage pattern that makes use of a binary bitmap, indicating non-zero values. Finally, the actual non-zero values are stored in a row-major order. This pruning technique maintains a model accuracy of 97.70\% while reaching an 87.00\% pruning ratio.

For FNM-Trans, by Zhang \textit{et al.}~\cite{Zhang:2024:FNM-Trans}, authors propose an accelerator that makes use of full $N\!:\!M$ Sparsity. $N\!:\!M$ sparsity is a structured sparsity technique where there are $N$ non-zero elements for every $M$ elements. FNM-Trans does apply $N\!:\!M$ sparsity to both the attention mechanism and weights. The authors propose a pruning algorithm to obtain high sparsity without impacting final model accuracy by a significant amount. The algorithm works in two stages. The first stage starts with $N = M$ for both attention and weights. Then, $N_a$ and $N_w$ are reduced with 1, followed by a model retraining using the new $N_a$ and $N_w$ parameters. This process of slowly reducing $N$ and subsequent trainings is continued until the model accuracy drops too much. The second stage is very similar to the first stage, except that instead of reducing both $N_a$ and $N_w$ at the same time, first $N_a$ is reduced until the minimum $N_a$ is reached. The same is done with $N_w$. The resulting model has up to 93.75\% attention sparsity and up to 75\% weight sparsity with only a 3.43\% reduction in model accuracy.

\section{IP design optimisations}
\label{sec:ip_design_optimisation}
Different computation and architecture designs that can be used for Transformer inference. Often, some form of Processing Elements (PE) are considered. Some implementations include designing dedicated PEs for specific calculations, whereas others employ generic PEs used for various purposes. Beyond PE designs, techniques such as data reuse or pipelining may be considered to increase throughput. Finally, various ways of increasing performance for computing non-linear functions, such as the SoftMax function, has been opted for.

\subsection{Computation: SoftMax and non-linear functions}
The SoftMax function has a significant performance impact on Transformer inference in FPGAs, due to its reliance on expensive operations. These operations, e.g., exponentiation and division, are highly inefficient on FPGA hardware. The frequent use of such operations in attention layers leads to an increase in computational and memory requirements, while its sequential nature limits parallelism and optimisation options.

A popular method of implementing the SoftMax function is through the use of Look-Up Tables (LUTs). For instance, Li~\cite{Li:2024:EEFA} has implemented such a SoftMax computation. However, the author notes that for a 16-bit SoftMax computation, the look-up table will be too large and will not fit onto the available memory. A more efficient SoftMax computation can be achieved by assuming a predefined threshold value and finding the maximum attention score. Then, for all other attention scores, if the difference with the maximum score is bigger than the threshold, it is discarded. In this manner, a minimum valid score is recorded and all valid entries between 1 and the threshold can be stored in a table. Each valid score is subtracted by the minimum value and is used in a look-up table. 

The authors of SWAT, Bai \textit{et al.}~\cite{Bai:2024:SWAT:arxiv}, note that standard implementations of Transformer models involve 3 distinct sequential steps: $\mathbf{Q} \mathbf{K}$ multiplication, SoftMax, and $\mathbf{S} \mathbf{V}$ multiplication. Often, each step is broken down into smaller tile-wise operations. However, since the SoftMax function is row-wise data dependent, it is not possible to compute it in a tiling window. This results in high number of memory transfers for loading and storing tile-wise intermediate results. Instead, the authors propose a kernel fusion optimisation technique. The SoftMax function is divided into two separate components: a numerator that does not depend on other elements of the row, and the denominator that depends on the sum of the exponential of all elements of the same row. In this fashion, the denominator can be placed after the $\mathbf{S} \mathbf{V}$ multiplication step, allowing the fusion of all three steps into a \enquote{unified row-wise kernel}. The restructuring of the SoftMax function~\cite{Bai:2024:SWAT:arxiv} is noted in mathematical terms in \Cref{eq:softmax_restructure}.
\begin{align}
	Z_{i,j} &= \sum_{n=0}^{H} S^\prime_{i,n} V_{n,j} 
	        = \sum_{n=0}^{H} \frac{\exp(S_{i,n})}{\sum_{l=0}^{H} \exp(S_{i,l})} V_{n,j} \notag\\
	        &= \frac{1}{\sum_{l=0}^{H} \exp(S_{i,l})} \sum_{n=0}^{H} \exp(S_{i,n}) V_{n,j}
	\label{eq:softmax_restructure}
\end{align}

A similar approach is followed in SALTS by Chen \textit{et al.}~\cite{Chen:2024:SALTS}, a flexible self-attention accelerator with Long Token Support. The authors note that the original SoftMax operation \enquote{requires traversing the input vector three times: element-wise exponent, reduced summation, and element-wise division}. By re-arranging the SoftMax computation, a fully pipelined design can be realised. Qin \textit{et al.}~\cite{Qin:2024:ELSI} have also employed the same technique.

These optimisations introduce important mathematical and hardware trade-offs, worth mentioning. Approximation techniques, such as threshold-based pruning, exploit the exponential decay of the SoftMax function, where elements significantly smaller than the maximum contribute very little to the final output. While this allows for substantial reductions in computation and memory usage, it can reduce accuracy in cases where attention distributions are relatively flat. Similarly, LUT-based or polynomial approximations introduce quantisation and approximation errors, which propagate non-linearly through the normalisation step of SoftMax. Prior work shows that such approximations can lower latency and power consumption, but at the cost of measurable numerical error, highlighting a clear trade-off between accuracy and efficiency~\cite{Leiva-Valverde:2025:QEAS, Vasyltsov:2021:ESAD}. In practice, carefully tuned approximations can often keep accuracy loss within 1\% for inference tasks, although sensitivity may increase for generative models or longer sequence lengths~\cite{Kim:2025:HAAS}.

Kernel fusion approaches, such as those proposed in SWAT~\cite{Bai:2024:SWAT:arxiv} and SALTS~\cite{Chen:2024:SALTS}, preserve the mathematical equivalence of the SoftMax operation while shifting the normalisation outside the matrix multiplication. Such restructuring reduces intermediate storage and off-chip memory traffic, which in turn enables streaming execution and better hardware utilisation. As SoftMax and related non-linear functions are major contributors to Transformer latency and its resource usage, such fusion techniques can significantly improve throughput and energy efficiency~\cite{Kim:2025:HAAS}. However, these listed benefits come at the cost of increased hardware complexity. In particular, inclusion of additional accumulators, wider data paths, and synchronisation logic, is needed to compute numerator and denominator terms concurrently. Fixed-point implementations must also carefully balance precision to avoid numerical instability during accumulation and scaling. Earlier FPGA-based studies further reveal that SoftMax implementations introduce non-trivial control and data path overhead. This is due to reduction operations and support for non-linear functions~\cite{Zhang:2020:EFIS}. What it boils down to is that these techniques trade modest accuracy loss and increased control complexity for substantial gains in memory efficiency and performance.

\subsection{Architecture designs: PE design}
A large majority of FPGA accelerator implementations employ \emph{Processing Elements (PEs)} as fundamental building blocks in their design. A PE is a design module dedicated to a specific computation or set of operations. Most works adopt architectures composed of multiple PEs, but differ in how computation is distributed amongst them. A common approach is the use of \emph{systolic arrays} consisting of many small, identical PEs, where each PE performs a simple, local operation and data flows regularly between neighbouring elements. Other implementations organise their architecture around multiple larger, functionally distinct PEs, with each PE responsible for a discrete computation stage or kernel. These architectural strategies are not necessarily mutually exclusive, and several designs combine elements of both to balance parallelism, resource utilisation, and design complexity. Furthermore, it is not always clearly distinguishable from the descriptions in the article which architectural elements are used in a given implementation. Based on the articles surveyed in this work, at least 25 employ PEs arranged in a systolic array, while at least 14 use architectures composed of several distinct PEs. Both collections are listed in \Cref{tab:processing_element_approach}.
\begin{table}[htbp]
	\centering
	\caption{List of articles taking advantage of different PE implementation approaches, which are not necessarily mutually exclusive.}
	\label{tab:processing_element_approach}
	\begin{tabular}{lc p{4.5cm}}
		\toprule
		\textbf{PE approach} & \textbf{Count} & \textbf{Articles} \\
		\midrule
		Systolic arrays 	& 25 	& \cite{Li:2024:EEFA, Marino:2024:ME-ViT:arxiv, Xiang:2025:ELIM:arxiv, Zhao:2025:HEAT, Shao:2024:FRAC, Du:2024:EEFB, Zhang:2024:FNM-Trans, Byun:2024:HFHR, Dong:2024:SWAT, Kabir:2024:RATN:arxiv, Zeng:2024:FlightLLM, Kabir:2024:ProTEA, Li:2024:MPTA, Dong:2025:UbiMoE, Liu:2025:FlightVGM, Zhang:2024:FVTA, Chen:2024:SALTS, Qiao:2025:COBRA:arxiv, Dong:2024:EQ-ViT, Qin:2024:ELSI, Sadeghi:2024:CHOSEN:arxiv, He:2025:F3, Moitra:2024:PIVOT, Kabir:2024:FAMOUS:arxiv, Sun:2025:DRViT} \\
		Distinct PEs 		& 14 	& \cite{Wang:2024:EFTA, Li:2024:EEFA, Zhao:2025:HEAT, Zhang:2024:FNM-Trans, Kabir:2024:ProTEA, Li:2024:MPTA, Dong:2025:UbiMoE, Liu:2025:FlightVGM, Chen:2024:SALTS, Dong:2024:EQ-ViT, Qin:2024:ELSI, He:2025:F3, Kabir:2024:FAMOUS:arxiv, Sun:2025:DRViT} \\
		\bottomrule
	\end{tabular}
\end{table}

An example of an implementation making use of different PE types is FAMOUS~\cite{Kabir:2024:FAMOUS:arxiv}. The architecture consist of 3 different modules, each with its own unique PE type. The \texttt{QKV} module is responsible for calculating the query, key and value matrices. The next module, called the \texttt{QK} module, is responsible for the matrix multiplication between the $\mathbf{Q}$ and $\mathbf{K}$ matrices and applying the SoftMax function. Finally, the \texttt{SV} module is responsible for multiplication with the Value matrix.

For FlightLLM by Zeng \textit{et al.}~\cite{Zeng:2024:FlightLLM}, a multi-core design has been employed, where each core includes a Matrix Processing Engine (MPE). Each MPE consists of multiple Matrix Processing Units that perform matrix calculations. The MPE supports two separate modes, an MM mode for Matrix-Matrix multiplication and an MV mode for Matrix-Vector multiplication. 

\subsection{Architecture designs: Pipelining}
Pipelining in FPGA design is an approach where operations are divided into separate stages, with each stage performing part of the computation. These stages are connected in sequence, each operating in parallel on different data during every clock cycle. Such a design increases throughput, as new data can enter the pipeline before the preceding data has finished processing, allowing for more efficient use of hardware resources. While in some architectures only a specific computation or part of the inference is pipelined~\cite{Kabir:2024:ProTEA}, other architectures are designed with full pipelining in mind~\cite{Guo:2024:HG-PIPE:arxiv}.

In the context of Transformer inference, pipelining introduces design challenges that go beyond generic throughput optimisation. First, attention-related computations contain inter-stage dependencies, especially when intermediate results must be reduced, normalised, or buffered before subsequent stages can proceed. Second, practical pipeline efficiency is often constrained not only by operator latency but also by memory bandwidth and data movement overhead, which can cause stalls even in deeply pipelined architectures. Third, workload characteristics such as sequence length, token granularity, and the distinction between encoder-style and autoregressive decoding workloads influence how effectively computation can be overlapped. As a result, successfully pipelined FPGA implementations of Transformers depend not only on inserting pipeline stages, but also on balancing dataflow, buffering, and memory access patterns across the attention and feed-forward blocks.

ProTEA~\cite{Kabir:2024:ProTEA} is an example of an architecture that does not necessarily implement a fully pipelined architecture, but does make use of pipelining within single computation steps. ProTEA has an architecture consisting of different PEs, targeting specific computations. The number of PEs depends on the unrolling factor and initiation interval of the pipelined loop.

\begin{figure}[htbp]
	\centering
	\includegraphics[width=\linewidth]{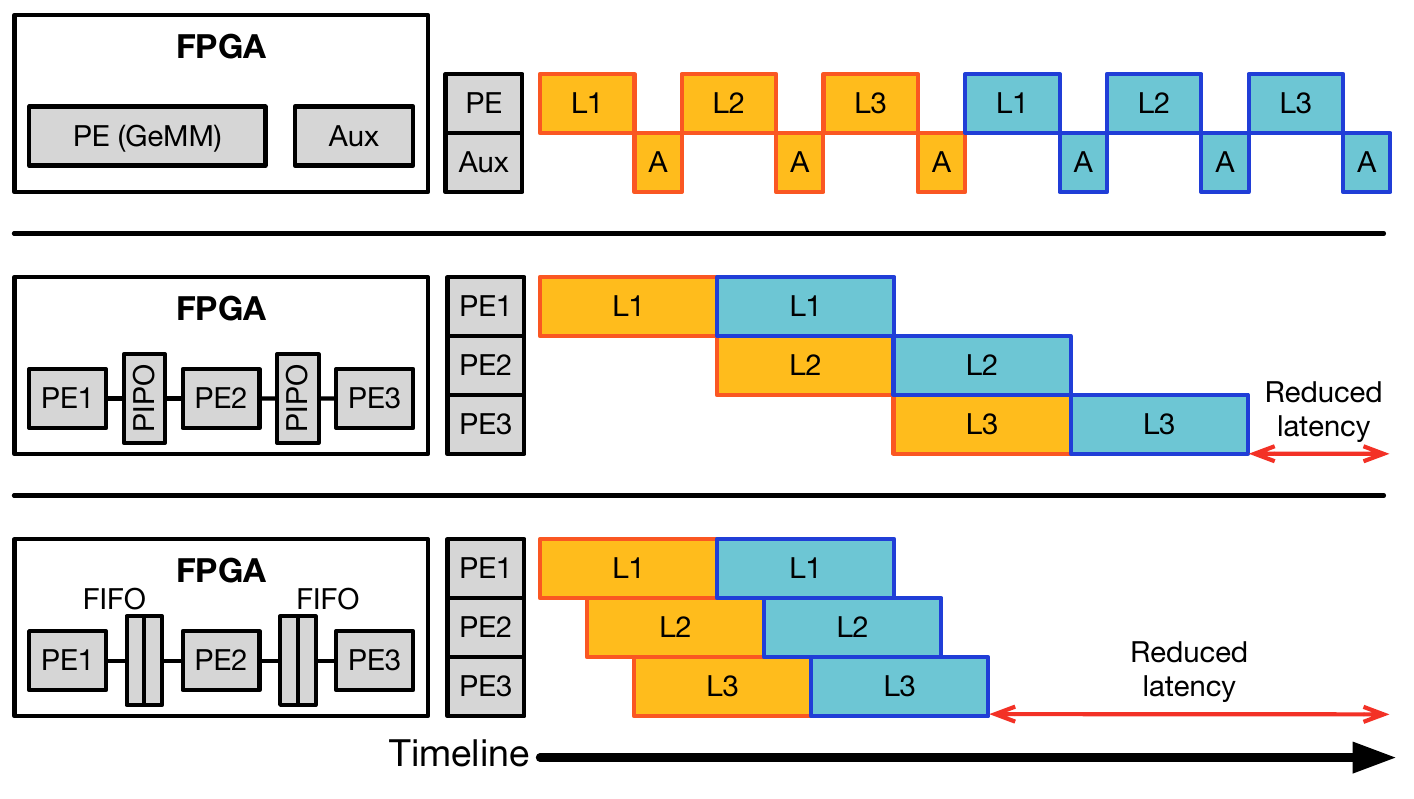}
	\caption{Three pipelining paradigms and resulting effects on the execution timeline, i.e., reduction in latency. From top to bottom: Temporal design, i.e., no pipelining; Coarse-grained pipelining; Fine-grained pipelining.}
	\label{fig:pipelining_pradigms}
\end{figure}

An instance of a fully pipelined architecture is HG-PIPE~\cite{Guo:2024:HG-PIPE:arxiv}. The architecture consist of a series of block levels, where each represents a single Encoder layer. Pipelining is performed both at the inter-block level and within each block. HG-PIPE employs a hybrid grained pipeline approach, i.e., while some computations can occur simultaneously, other computations are bound to start once previous steps are complete. A diagram describing the execution impact of different pipelining paradigms is provided in \Cref{fig:pipelining_pradigms} (inspired by~\cite{Guo:2024:HG-PIPE:arxiv}). The incorporated elements in addition to PE are: General Matrix-Matrix Multiplication (GeMM); Parallel-In Parallel-Out (PIPO), which is a type high-bandwidth parallel data register; Auxiliary (Aux) as supporting or helper hardware modules, e.g., non-linear functions; and First-In First-Out (FIFO), which is used for ordered buffering to achieve data streaming between accelerator stages.

Overall, a large majority of articles mention the use of pipelining in some form, either within single computation steps or on a larger architectural scale. In total, 68\% of the collected articles explicitly mention the use of pipelining. The remaining 32\% do not make any comments regarding pipelining.

\subsection{Architecture designs: Instruction Set Architectures}
For this design, FlightLLM~\cite{Zeng:2024:FlightLLM} and FlightVGM~\cite{Liu:2025:FlightVGM} are two accelerators that make use of a custom Instruction Set Architecture (ISA) between the host and the hardware accelerator. The ISA consists of 6 instructions. First, the Load (LD) and Store (ST) instructions covered, which transfer data between off-chip memory and on-chip buffers. The Matrix-Matrix multiplication (MM) and Matrix-Vector multiplication (MV) instructions are added for matrix multiplication operations. The MISC instruction is included for Layer Normalisation and SoftMax calculations. Finally, the SYS instruction covers the synchronisation between the host CPU and multiple Super Logic Regions (SLRs), after each inference execution is completed.

\section{Insights and discussion}
\label{sec:insights}
Given what we have gathered from the pool of articles in this survey, there are comparisons and analyses worthy of sharing. When it comes to performance comparison, the two relatively common metrics to rely on are: \textit{Throughput in GOPS} and \textit{Energy efficiency in GOPS/Watts}.

Reported hardware metrics (GOPS and GOPS/W) should be interpreted with care when comparing across different articles. These values are influenced not only by the accelerator architecture itself, but also by the underlying model class, numerical precision, sparsity pattern, target FPGA platform, and evaluation methodology. In particular, comparisons across encoder-oriented workloads, generative decoding workloads, and highly quantised designs are not strictly uniform. Accordingly, the following analysis is intended to highlight broad tendencies rather than establish definitive rankings across fundamentally different implementations.

\subsection{Performance comparison: Throughput}
Focusing on throughput as a performance metric, \Cref{tab:top_ten_troughput} lists the top 10 implementations with the highest throughput in Giga Operations Per Second (GOPS), sorted from high to low. GOPS is used to measure how many billion operations can be performed per second. In the context of Transformer machine learning inference on FPGAs, GOPS gives an idea of the raw compute throughput achieved when running inference workloads. As can be seen in \Cref{tab:top_ten_troughput}, HG-PIPE~\cite{Guo:2024:HG-PIPE:arxiv} has by far the highest throughput, with a throughput more than twice as much as the second best implementation by Liu \textit{et al.}~\cite{He:2025:F3}.
\begin{table*}[htbp]
	\centering
	\caption{Top 10 implementations with the highest throughput in GOPS, sorted from high to low. Note that not every field is reported in every article. Cells designated with \enquote{n/a} indicate data that were unavailable or not reported in the original paper.}
	\label{tab:top_ten_troughput}
	\begin{tabular}{l l c c c c c l | >{\centering\arraybackslash}p{2cm} | c}
		\toprule
		\textbf{Design level} & 
		\textbf{Storage} & 
		\textbf{FF} & 
		\textbf{LUT} & 
		\textbf{DSP} & 
		\textbf{BRAM} & 
		\textbf{Bit-width} & 
		\textbf{Innovation} & 
		\textbf{Throughput\newline (GOPS)} & 
		\textbf{Article} \\
		\midrule
		HLS 	& On-chip 	& n/a 		& 669K 	& 312 		& 1\,006 	& 3 	& LUT based MAC 			& \textbf{17\,795} 		& \cite{Guo:2024:HG-PIPE:arxiv} \\
		RTL 	& Off-chip 	& 578K 		& 472K 	& 9\,024 	& 1\,520 	& n/a 	& Flexible floating point 	& \textbf{8\,204} 		& \cite{He:2025:F3} \\
		HLS 	& Off-chip 	& 162K 		& 122K 	& 78 		& 691.5 	& 1 	& Binary weights 			& \textbf{3\,894.74} 	& \cite{Qiao:2025:COBRA:arxiv} \\
		n/a 	& Off-chip 	& 1507K 	& 557K 	& 5\,096 	& 1\,581 	& 16 	& $N\!:\!M$ sparsity 		& \textbf{2\,750} 		& \cite{Zhang:2024:FNM-Trans} \\
		n/a 	& On-chip 	& 139K 		& 118K 	& 2\,147 	& 283 		& 8 	& Weight loop dataflow 	& \textbf{2\,330.2} 	& \cite{Zhang:2024:FVTA} \\
		HLS 	& Off-chip 	& n/a 		& n/a 	& 3\,482 	& 1\,162 	& 16 	& Attention fusion 			& \textbf{1\,420} 		& \cite{Qin:2024:ELSI} \\
		n/a 	& On-chip 	& 548K 		& 274K 	& 2\,520 	& 912 		& 1 	& Binary weights 			& \textbf{1\,387.59} 	& \cite{Ji:2024:BETA} \\
		RTL 	& Off-chip 	& n/a 		& 493K 	& 5\,016 	& n/a 		& 8, 4 	& Sorting engine 			& \textbf{1\,223.9} 	& \cite{Wang:2024:TransFRU} \\
		RTL 	& Off-chip 	& 114K 		& 128K 	& 768 		& n/a 		& 5 	& Hybrid quantisation 		& \textbf{1\,023.3} 	& \cite{Zhao:2025:HEAT} \\
		n/a 	& On-chip 	& 1083K 	& 544K 	& 10\,812 	& 1\,903 	& 8 	& Redundancy aware 			& \textbf{986.3} 		& \cite{Sun:2025:DRViT} \\
		\bottomrule
	\end{tabular}
\end{table*}

HG-PIPE also has the highest number of consumed LUTs, which is expected as the implementation relies on LUT-based computations. Interestingly, while \Cref{tab:weights_storage_approach_comparison} indicates that implementations utilising on-chip memory have on average a much higher throughput, which is also a loose theoretical expectation, this does not hold. Considering the top 10 highest throughput implementations, only 4 take advantage of on-chip storage, proving that it is still possible to achieve relatively equal performance when using off-chip memory. However, we must note that the top contender, HG-PIPE, is based on on-chip storage and it achieves a considerably higher throughput compared to the 2nd best implementation.

\subsection{Performance comparison: Energy efficiency}
Deducing energy efficiency is less straightforward as it depends on, simply put, the amount of work being done, which translates to data throughput per platform at hand. When comparing implementations, it is tempting to focus only on raw performance metrics, such as throughput and power. However, these metrics both are highly dependent on the implemented model itself, as well as the size of the FPGA, i.e., both are use-case and experiment specific. Therefore, energy efficiency (GOPS/Watts), i.e., the amount of energy spent per unit of throughput, can be an additional metric to consider for a fair comparison. Similar to our throughput metric listing, \Cref{tab:top_ten_efficiency} lists the top 10 implementations with the best energy efficiency, sorted from high to low. This selection is limited to those articles reporting both power usage and throughput. As it can be seen, the top two implementations both use binarised quantisation, which allows for very fast and efficient computations. However, switching to binarised weights is not always possible for all models. Furthermore, a considerable number of implementations (5 articles, half) within this top 10 batch, use RTL-level design instead of HLS, much higher than the average use of RTL amongst all articles. According to \Cref{tab:fpga_tooling}, the average use of RTL is about 30\%. Although, these ratios are to be taken with a grain of salt, since a considerable number of articles, seen in \Cref{tab:fpga_tooling}, do not specify the choice of design level.
\begin{table*}[htbp]
	\centering
	\caption{Top 10 implementations with the highest power efficiency (throughput/power), sorted from high to low. Note that not every field is reported in every article. Cells designated with \enquote{n/a} indicate data that were unavailable or not reported in the original paper.}
	\label{tab:top_ten_efficiency}
	\begin{tabular}{l c c c c c l | c | c}
		\toprule
		\textbf{Design level} &
		\textbf{FF} &
		\textbf{LUT} &
		\textbf{DSP} &
		\textbf{BRAM} &
		\textbf{Bit-width} &
		\textbf{Innovation} &
		\textbf{Efficiency (GOPS/W)} &
		\textbf{Article} \\
		\midrule
		RTL 	& 73K 		& 110K 	& 0 		& 177 		& 1 	& Binary weights 				& \textbf{679.76} (944.87 / 1.39)		& \cite{Du:2024:EEFB} \\
		HLS 	& 162K 		& 122K 	& 78 		& 691.5 	& 1 	& Binary weights 				& \textbf{448.70} (3\,894.74 / 8.68) 	& \cite{Qiao:2025:COBRA:arxiv} \\
		HLS 	& n/a 		& 669K 	& 312 		& 1\,006 	& 3 	& LUT based MAC 				& \textbf{381.05} (17\,795 / 46.7)		& \cite{Guo:2024:HG-PIPE:arxiv} \\
		RTL 	& 578K 		& 472K 	& 9\,024 	& 1\,520 	& n/a 	& Flexible floating point 		& \textbf{189.91} (8\,204 / 43.2)		& \cite{He:2025:F3} \\
		n/a 	& 548K 		& 274K 	& 2\,520 	& 912 		& 1 	& Binary weights 				& \textbf{174.54} (1\,387.59 / 7.95)	& \cite{Ji:2024:BETA} \\
		n/a 	& 139K 		& 118K 	& 2\,147 	& 283 		& 8 	& Weight loop dataflow 		& \textbf{109.04} (2\,330.2 / 21.37)	& \cite{Zhang:2024:FVTA} \\
		RTL 	& n/a 		& n/a 	& 1\,024 	& n/a 		& 8 	& Intra-/inter-layer fusions 	& \textbf{104.98} (780 / 7.43)			& \cite{Shao:2024:FRAC} \\
		n/a 	& 1\,507K 	& 557K 	& 5\,096 	& 1\,581 	& 16 	& $N\!:\!M$ sparsity 			& \textbf{97.41} (2\,750 / 28.23)		& \cite{Zhang:2024:FNM-Trans} \\
		RTL 	& 114K 		& 128K 	& 768 		& n/a 		& 5 	& Hybrid quantisation 			& \textbf{43.58} (1\,023.3 / 23.48)	& \cite{Zhao:2025:HEAT} \\
		RTL 	& n/a 		& n/a 	& n/a 		& n/a 		& 8 	& n/a 							& \textbf{39.10} (385.5 / 9.86)		& \cite{Chen:2024:SALTS} \\
		\bottomrule
	\end{tabular}
\end{table*}

\subsection{Impact on model performance}
Yet another important aspect to consider is the actual impact of applied innovation, the accelerator implementation itself, or the combination of both, on model accuracy. Inline with discussions on throughput and energy efficiency, we consider the union of articles from \Cref{tab:top_ten_troughput,tab:top_ten_efficiency} as our pool of articles.

One limitation to note is that not every article reports the accuracy impact using a consistent definition. Some articles take an existing model, quantise and accelerate it, and then compare the accuracy with that of the original model. Other articles take an already quantised model, for instance a binary quantised model, accelerate it, and then report the resulting accuracy loss. In the latter case, the reported accuracy loss is primarily due to the accelerator implementation rather than the impact of the quantisation technique itself. In short, there is no consistency amongst the articles when it comes to picking a baseline to compare against. Given the extracted information, performing a robust comparative analysis of innovation impact is not practically feasible.

As quantisation is a key technique in achieving light-weight model variants, suitable for acceleration, it must also be noted that the chosen type of quantisation defines not only the impact, but also the amount of effort to realise it. Examples such as, Post-Training Quantisation (PTQ), Quantisation-Aware Training (QAT), weight-only quantisation, mixed-precision quantisation, binary or ternary quantisation, and so forth, each demand different levels of access to models and impose different computational requirements. It is conceivable that such imposed requirements is a motivation, or rather a deterring factor, for researchers to consider already quantised models as their implementation baseline.

We list the combined impact (resulting from the innovation applied and implementation performed) on model output prediction performance where it is provided, in \Cref{tab:accuracy_impact}. The table covers high-level model types, considering the introduced types in \Cref{sec:background}, as well as specific model variants. Needless to say, sensitivity to quantisation, even if it is the same technique, is highly model and data dependant.
\begin{table*}[htbp]
	\centering
	\caption{The combined impact of implementations on model output prediction accuracy, for articles of the top 10 lists, alongside relevant model types and specific model variants per article. Cells designated with \enquote{n/a} indicate data that were unavailable or not reported in the original paper.}
	\label{tab:accuracy_impact}
	\begin{tabular}{l l l | l | c}
		\toprule
		\textbf{Innovation} &
        \textbf{Model type} &
		\textbf{Model variant} &
		\textbf{Accuracy impact} &
        \textbf{Article} \\
		\midrule
		LUT based MAC              & Vision Transformer                   & DeiT-tiny             & 74.5\% $\rightarrow$ 71.05\%  & \cite{Guo:2024:HG-PIPE:arxiv} \\
		Weight loop dataflow       & Vision Transformer                   & Various ViT           & None                          & \cite{Zhang:2024:FVTA} \\
		Hybrid quantisation        & Vision Transformer                   & Swin-T                & 0.526\% loss                  & \cite{Zhao:2025:HEAT} \\
		Redundancy aware           & Vision Transformer                   & ViT-Tiny              & n/a                           & \cite{Sun:2025:DRViT} \\
		Intra-/inter-layer fusion  & Vision Transformer                   & EfficientViT          & n/a                           & \cite{Shao:2024:FRAC} \\
        \midrule
		Flexible floating point    & Language model                       & BERT-small/DistilBERT & $\leq$ 1\% loss               & \cite{He:2025:F3} \\
		Binary weights             & Language model                       & BERT-base             & 83.9\% $\rightarrow$ 68.2\%   & \cite{Qiao:2025:COBRA:arxiv} \\
		$N\!:\!M$ sparsity         & Language model                       & BERT-base             & 3.43\% loss                   & \cite{Zhang:2024:FNM-Trans} \\
		Attention fusion           & Language model                       & BERT-base             & None                          & \cite{Qin:2024:ELSI} \\
		Binary weights             & Language model                       & BinaryBERT            & None                          & \cite{Ji:2024:BETA} \\
		Sorting engine             & Language model                       & BERT-base             & n/a                           & \cite{Wang:2024:TransFRU} \\
        \midrule
		n/a                        & Language model / Vision Transformer  & BERT-base / ViT       & $\sim$0.5--0.65\% loss        & \cite{Chen:2024:SALTS} \\
        \midrule
		Binary weights             & n/a                                  & n/a                   & None                          & \cite{Du:2024:EEFB} \\
		\bottomrule
	\end{tabular}
\end{table*}

\subsection{Effective implementations}
\begin{figure}[htbp]
	\centering
	\includegraphics[width=\linewidth]{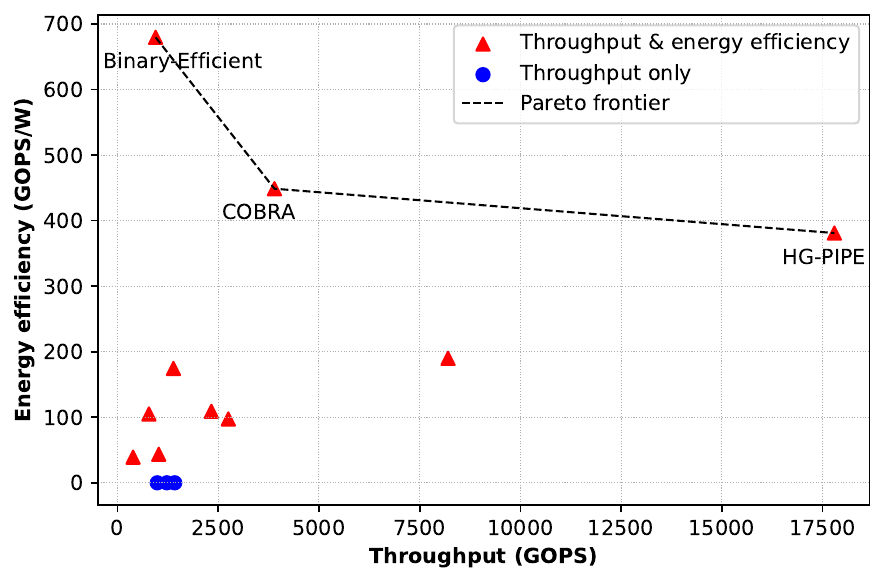}
	\caption{Throughput versus energy efficiency for FPGA-based inference accelerators, designating the Pareto-optimal designs and the Pareto frontier.}
	\label{fig:throughput_energy_efficiency_pareto}
\end{figure}

\Cref{fig:throughput_energy_efficiency_pareto} plots throughput versus energy efficiency for FPGA-based inference accelerators listed in \Cref{tab:top_ten_troughput,tab:top_ten_efficiency}. Designs with both metrics available are shown with red markers, while designs reporting only throughput are shown with blue markers. Note that not every implementation provides both metrics and there is an intersection between the two tables. The dashed line denotes the Pareto frontier, representing designs for which no other design achieves both higher throughput and higher energy efficiency, simultaneously. The plot highlights the trade-off between raw performance and energy efficiency and as we can see, a small number of Pareto-optimal designs best balance these competing objectives.

\begin{figure}[htbp]
	\centering
	\includegraphics[width=\linewidth]{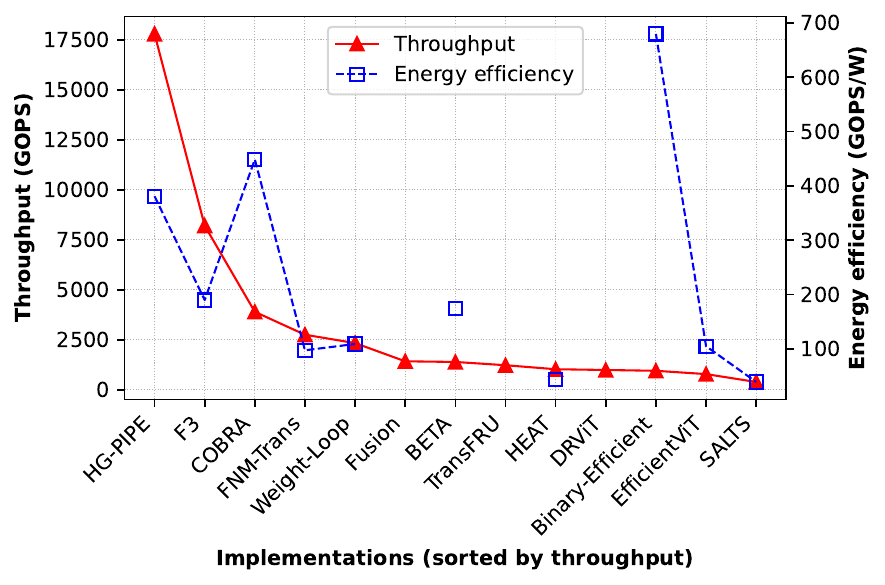}
	\caption{Energy efficiency trends for FPGA-based inference accelerators, with respect to ranked throughput.}
	\label{fig:throughput_energy_efficiency_ranked}
	\vspace{0.25em}
    \begin{minipage}{\linewidth}
    	\footnotesize
    	\raggedright
		\textbf{Legend.}
        HG-PIPE~\cite{Guo:2024:HG-PIPE:arxiv},
        F3~\cite{He:2025:F3},
        COBRA~\cite{Qiao:2025:COBRA:arxiv},
        FNM-Trans~\cite{Zhang:2024:FNM-Trans},
        Weight-Loop~\cite{Zhang:2024:FVTA},
        Fusion~\cite{Qin:2024:ELSI},
        BETA~\cite{Ji:2024:BETA},
        TransFRU~\cite{Wang:2024:TransFRU},
        HEAT~\cite{Zhao:2025:HEAT},
        DRViT~\cite{Sun:2025:DRViT},
        Binary-Efficient~\cite{Du:2024:EEFB},
        EfficientViT~\cite{Shao:2024:FRAC},
        SALTS~\cite{Chen:2024:SALTS}.
    \end{minipage}
\end{figure}

As an alternative perspective, \Cref{fig:throughput_energy_efficiency_ranked} shows how energy efficiency evolves as we move from the highest-throughput designs to the lower-throughput ones. Here, throughput-sorted designs are given with throughput and energy efficiency plotted on separate axes. While throughput decreases monotonically by construction, energy efficiency exhibits non-monotonic behaviour, indicating that high throughput does not necessarily imply poor energy efficiency. This further motivates the use of a Pareto-based analysis.

Note that while such plots are revealing, it is hard to arrive at any solid conclusion. As little generality can be claimed beyond applied methodologies for implementations targeting FPGAs, such analyses are to be taken with a grain of salt. After all, a tremendous amount of customisation goes into each individual implementation and these custom techniques are not common amongst the implementations.

\subsection{Standardised model benchmarking}
One of the major challenges in conducting surveys on this topic is the difficulty of making meaningful comparisons between different accelerator implementations. The primary reason is the inconsistent choice of Transformer models, or different variants of the same model, to be evaluated, making fair comparisons largely infeasible. The chosen model has a significant impact on performance, resource utilisation, and energy efficiency of accelerator implementations.

Several models are frequently used in the literature, such as BERT~\cite{Devlin:2019:BERT}, Swin~\cite{Liu:2021:HVTS}, and DeiT~\cite{Touvron:2021:TDEI}. However, these models represent families of architectures rather than fixed specifications and exist in numerous configurations, differing in structure and parameter counts. Consequently, reported performance figures often reflect model complexity just as much as architectural or algorithmic innovation.

Using one or more standard Transformer benchmark models would make future work easier to compare by standardising the model structure and workload. In addition, a common benchmarking approach with common sets of performance metrics, measurement conditions, and reporting practices, will further reduce uncertainty and improve reproducibility. It is important to ensure that reported performance gains are primarily resulting from accelerator design choices, rather than differences in model variants or evaluation methods. While standard models and benchmarking practices would not replace application-specific evaluations, they could serve as a common reference point. At a minimum, consistently reporting the evaluated model's size, configuration, parameter count, and benchmarking setup, will provide valuable context and enable more meaningful comparisons.

\section{Conclusion}
\label{sec:conclusion}
We have provided a comprehensive and systematic review of recent articles on the topic of \emph{Transformer model inference deployment on FPGA platforms}. We have followed a repeatable protocol for collection and pruning of articles, which also renders the review easily expandable by addition of new articles. The extracted taxonomy, \Cref{fig:topics_taxonomy}, depicts a high-level, but complete hierarchy of the focus area within the literature. The review culminates in top-10 listings based on the two main performance metrics commonly reported, throughout (GOPS) and power (W), from which we derive power efficiency (GOPS/W) as a more meaningful ranking metric.

We observe that there is no absolute preference across any of the examined categorical dimensions. Whether considering design abstraction level (HLS vs. RTL), memory organisation (off-chip vs. on-chip), pipelining strategies, or quantisation schemes, the choices remain rather implementation-specific and application-driven. While this diversity highlights the inherent flexibility of FPGA platforms, it also underscores the difficulty of drawing direct comparisons across published works. In particular, we have touched upon the lack of standardised benchmarking methodologies, as well as incomplete reporting of architectural and implementation details, as a main challenge to comparisons. 

Quantisation deserves special emphasis as a double-edged sword. While reduced precision and lower bit-width arithmetic can yield substantial gains in performance and energy efficiency, they may also incur non-negligible degradation in model accuracy, to the point that some ML models will no longer be viable if quantised. Addressing this trade-off, alongside improved reporting practices and benchmark standardisation, represents a key opportunity for future work in FPGA-based Transformer inference.

A final and ever-present challenge lies in the trade-off between architectural flexibility and peak performance. Highly specialised FPGA accelerators, tailored to a specific Transformer model or configuration, can achieve substantial gains in throughput and energy efficiency, but at the cost of limited support for alternative models or workloads. In contrast, more flexible designs that accommodate a wider range of model variants and parameters typically incur additional overheads, reducing raw performance. This trade-off is inherent rather than incidental: designs optimised for maximum performance generally sacrifice flexibility, while general-purpose accelerators must accept lower peak efficiency. Consequently, the appropriate balance remains application-dependent and driven by deployment requirements.

\begin{acks}
This publication is part of the project ZORRO with project number KICH1.ST02.21.003 of the research programme Key Enabling Technologies (KIC), which is (partly) financed by the Dutch Research Council (NWO).
\end{acks}
